\documentclass[acmtog,authorversion,nonacm]{acmart}
\acmSubmissionID{227}

\makeatletter
\let\@authorsaddresses\@empty
\makeatother

\usepackage{booktabs} %
\usepackage{bold-extra}

\usepackage[group-separator={,}]{siunitx}

\acmJournal{TOG}

\usepackage{soul}

\newcommand{\modelname}{Camera Preconditioning\xspace}
\newcommand{\nameacronym}{CamP\xspace}
\newcommand{\namehandle}{\textbf{\nameacronym}\xspace}

\usepackage{mathtools}
\usepackage{bm}
\usepackage{xfrac}
\usepackage{xspace}
\usepackage[makeroom]{cancel}
\usepackage{makecell}
\usepackage{multirow}

\usepackage{subcaption}
\usepackage{color}
\usepackage{xcolor}
\usepackage{colortbl}
\usepackage{lipsum}
\usepackage{wasysym }

\usepackage[abs]{overpic}

\definecolor{myyellow}{rgb}{1,1, 0.6}
\definecolor{myorange}{rgb}{1, 0.8, 0.6}
\definecolor{myred}{rgb}{1, 0.6, 0.6}
\definecolor{newcolor}{HTML}{3f37c9}

\newcommand\rurl[1]{%
  \href{https://#1}{\nolinkurl{#1}}%
}

\DeclareMathOperator{\diag}{diag}

\newcommand{\updel}{\mathop{}\!\upDelta}

\newcommand{\figref}[1]{Fig.~\ref{#1}}
\newcommand{\tabref}[1]{Tab.~\ref{#1}}

\newcommand{\mat}[1]{\bm{\mathrm{#1}}}

\newcommand{\sethree}[0]{\mathfrak{se}(3)}

\newcommand{\SOTHREE}[0]{\text{SO}(3)}
\newcommand{\logrot}[0]{\mat{r}}

\def\figcell#1#2#3{\begin{subfigure}{#1\columnwidth}\centering\includegraphics[width=\textwidth]{#2}
\def\temp{#3}\ifx\temp\empty\else\caption*{\centering{#3}}\fi\end{subfigure}}

\def\figcellc#1#2#3{\begin{subfigure}{#1\columnwidth}\centering\includegraphics[width=\textwidth]{#2} \def\temp{#3}\ifx\temp\empty\else\caption{#3}\fi\end{subfigure}}

\def\figcellt#1#2#3#4{\begin{subfigure}{#1\columnwidth}\centering\includegraphics[width=\textwidth,#4]{#2} \def\temp{#3}\ifx\temp\empty\else\caption*{#3}\fi\end{subfigure}}

\newcommand{\figcellbox}[1]{\fcolorbox{black}{white}{#1}}

\newcommand{\figcellboxwhite}[1]{\fcolorbox{white}{white}{#1}}

\def\figcelltb#1#2#3#4{\begin{subfigure}{#1\columnwidth}\centering\figcellbox{\includegraphics[width=\textwidth,#4]{#2}} \def\temp{#3}\ifx\temp\empty\else\caption*{#3}\fi\end{subfigure}}

\def\figcelltbwhite#1#2#3#4{\begin{subfigure}{#1\columnwidth}\centering\figcellboxwhite{\includegraphics[width=\textwidth,#4]{#2}} \def\temp{#3}\ifx\temp\empty\else\caption*{#3}\fi\end{subfigure}}

\def\figcellb#1#2#3{\begin{subfigure}{#1\columnwidth}\centering\figcellbox{\includegraphics[width=\textwidth]{#2}}
\def\temp{#3}\ifx\temp\empty\else\caption*{\centering{#3}}\fi\end{subfigure}}

\makeatletter
\newcommand{\thickhline}{%
    \noalign {\ifnum 0=`}\fi \hrule height 1pt
    \futurelet \reserved@a \@xhline
}
\newcolumntype{"}{@{\hskip\tabcolsep\vrule width 1pt\hskip\tabcolsep}}
\makeatother

\usepackage{booktabs}

\DeclareGraphicsExtensions{.png,.jpg,.pdf,.ai,.psd}
\newcommand{\onehalf}{\sfrac{1}{2}}

\def\eg{\emph{e.g.}\@\xspace}

\makeatletter
\newcommand\footnoteref[1]{\protected@xdef\@thefnmark{\ref{#1}}\@footnotemark}
\makeatother

\newcommand{\camerafont}[1]{\textsf{\textbf{#1}}}
\newcommand{\camerafontnobf}[1]{\textsf{#1}}

\def\varpixel{\mat{p}}

\def\varpoint{\mat{x}}

\def\varprojection{\Pi}

\def\varcamparam{\mat{\phi}}
\def\vardeltacamparam{\updel \mat{\phi}}
\def\varmodelparam{\mat{\theta}}

\def\varjac{\mat{J}}

\def\varprecon{\mat{P}^{-1}}

\def\varcov{\mat{\Sigma}}

\def\varRthree{\mathbb{R}^3}
\def\varRtwo{\mathbb{R}^2}
\def\varR{\mathbb{R}}
\def\varmodel{M}
\def\varloss{\mathcal{L}}
\def\varpixelvalue{\mat{c}}

\begin{document}

\title{\nameacronym: \modelname for Neural Radiance Fields}

\author{Keunhong Park}
\affiliation{%
 \institution{Google Research}
 \city{San Francisco    }
 \state{CA}
 \postcode{94105}
 \country{USA}
}
\email{keunhong@google.com}

\author{Philipp Henzler}
\affiliation{%
 \institution{Google Research}
 \city{San Francisco}
 \state{CA}
 \postcode{94105}
 \country{USA}
}
\email{phenzler@google.com}

\author{Ben Mildenhall}
\affiliation{%
 \institution{Google Research}
 \city{San Francisco}
 \state{CA}
 \postcode{94105}
 \country{USA}
}
\email{bmild@google.com}

\author{Jonathan T. Barron}
\affiliation{%
 \institution{Google Research}
 \city{San Francisco}
 \state{CA}
 \postcode{94105}
 \country{USA}
}
\email{barron@google.com}

\author{Ricardo Martin-Brualla}
\affiliation{%
 \institution{Google Research}
 \city{Seattle}
 \state{CA}
 \postcode{98103}
 \country{USA}
}
\email{rmbrualla@google.com}

\renewcommand\shortauthors{Park, K. et al}

\begin{abstract}
Neural Radiance Fields (NeRF) can be optimized to obtain high-fidelity 3D scene reconstructions of objects and large-scale scenes. 
However, NeRFs require accurate camera parameters as input --- inaccurate camera parameters result in blurry renderings. 
Extrinsic and intrinsic camera parameters are usually estimated using Structure-from-Motion (SfM) methods as a pre-processing step to NeRF, but these techniques rarely yield perfect estimates. 
Thus, prior works have proposed jointly optimizing camera parameters alongside a NeRF, but these methods are prone to local minima in challenging settings.
In this work, we analyze how different camera parameterizations affect this joint optimization problem, and observe that standard parameterizations exhibit large differences in magnitude with respect to small perturbations, which can lead to an ill-conditioned optimization problem.
We propose using a proxy problem to compute a whitening transform that eliminates the correlation between camera parameters and normalizes their effects, and we propose to use this transform as a \emph{preconditioner} for the camera parameters during joint optimization. 
Our preconditioned camera optimization significantly improves reconstruction quality on scenes from the Mip-NeRF 360 dataset: we reduce error rates (RMSE) by 67\% compared to state-of-the-art NeRF approaches that do not optimize for cameras like Zip-NeRF, and by 29\% relative to state-of-the-art joint optimization approaches using the camera parameterization of SCNeRF.
Our approach is easy to implement, does not significantly increase runtime, can be applied to a wide variety of camera parameterizations, and can straightforwardly be incorporated into other NeRF-like models.

\end{abstract}



\newcommand{\macrowidth}{0.98in}

\begin{teaserfigure}
    \centering
    \begin{tabular}{@{}c@{\,}c@{\,}c@{\,}c@{\,}c@{}}
        \includegraphics[width=3.04in]{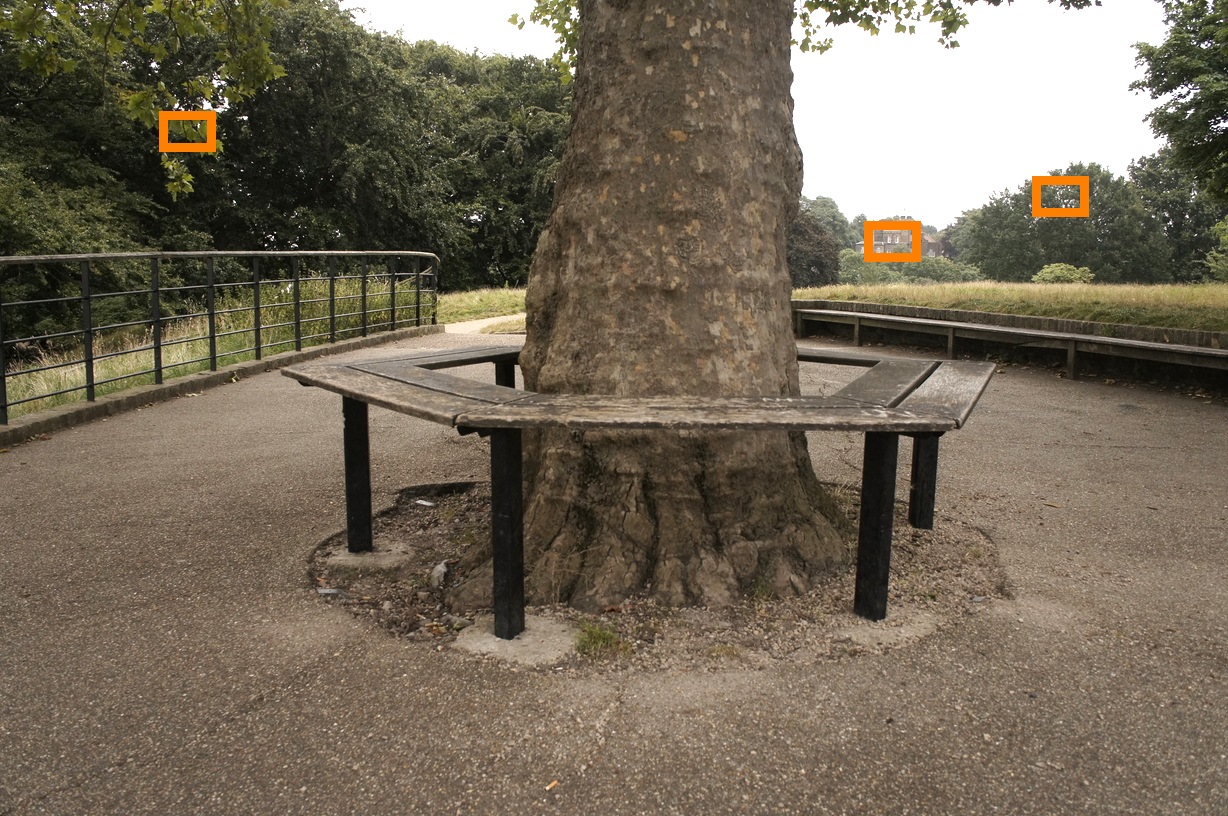} &
        \includegraphics[width=\macrowidth]{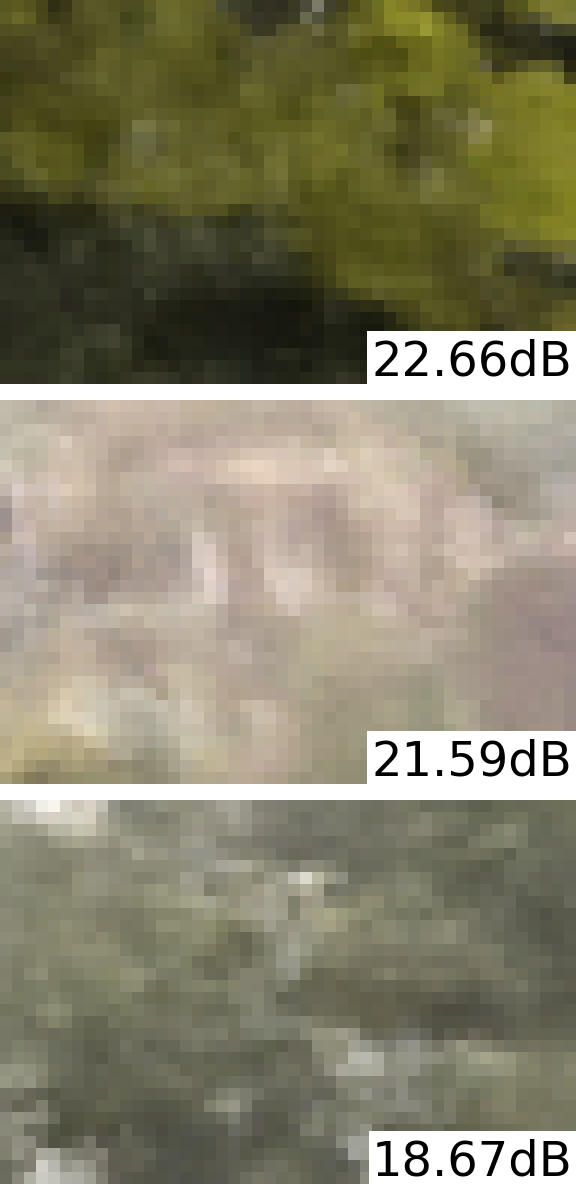} & 
    	\includegraphics[width=\macrowidth]{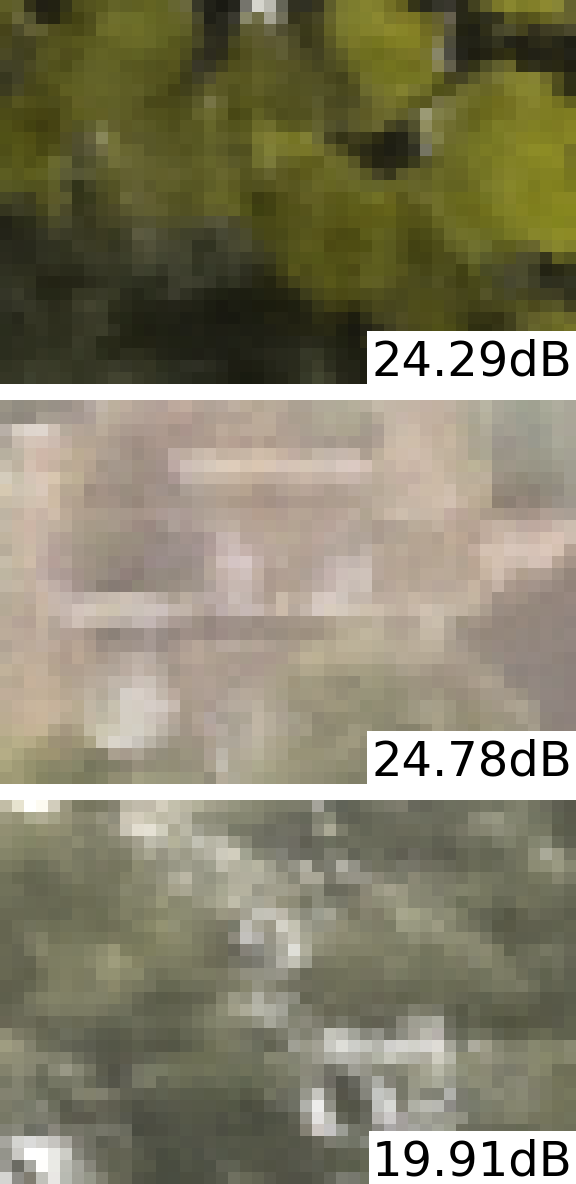} &
    	\includegraphics[width=\macrowidth]{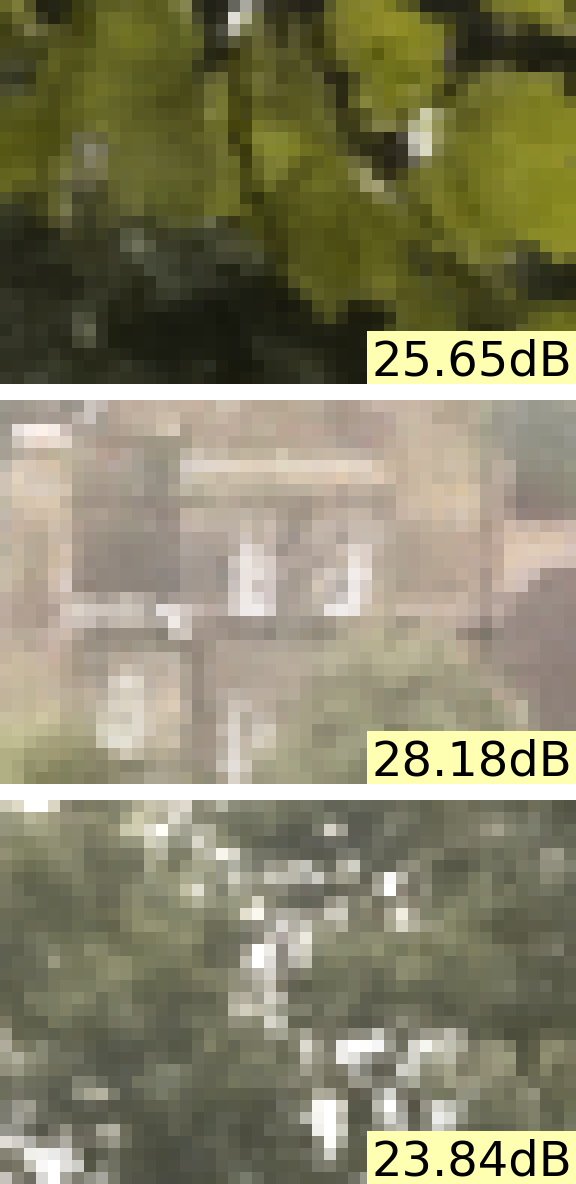} &
    	\includegraphics[width=\macrowidth]{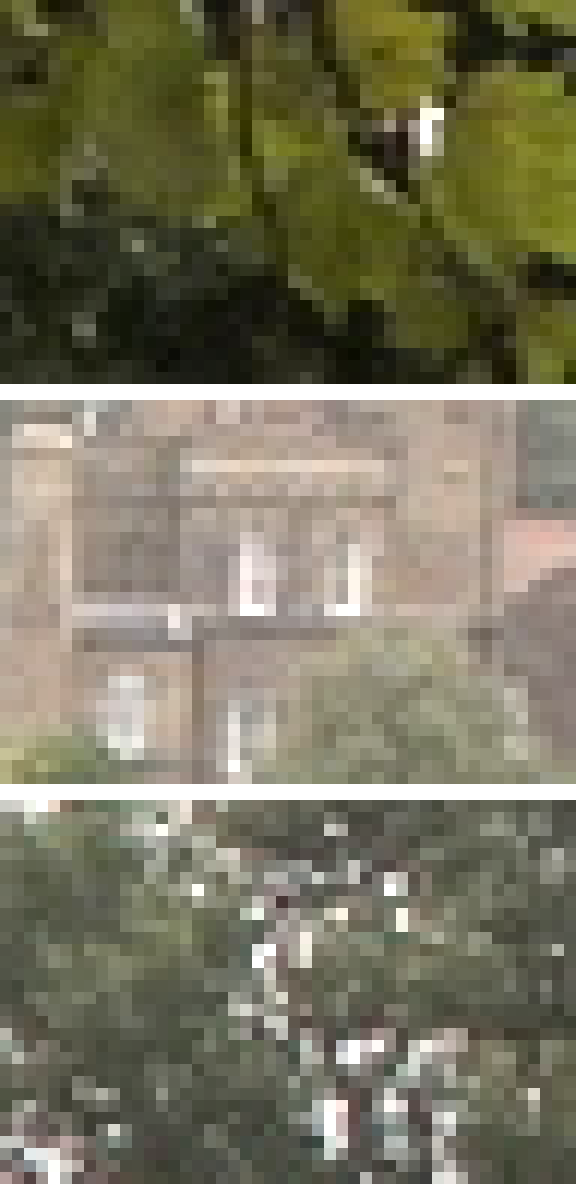} \\
    	\small (a) Ground Truth Image, with Annotations &
        \small (b) Zip-NeRF  &
        \small (c) +Camera Optim. & 
        \small \!\!\!\! (d) +\bf{\nameacronym}\!\! & 
        \small (e) Ground Truth \\
    \end{tabular}
    \caption{
    Our simple method preconditions camera optimization in camera-optimizing Neural Radiance Fields, significantly improving their ability to jointly recover the scene and camera parameters. (b) We apply our method to Zip-NeRF~\cite{barron2023zip}, a state-of-the-art NeRF approach that relies entirely on camera poses obtained from COLMAP~\cite{schoenberger2016sfm}. (c) Adding joint camera optimization to Zip-NeRF (using the camera parameterization of SCNeRF~\cite{jeong2021self}) improves image quality in distant parts of the scene, but only modestly. (d) Our proposed  \emph{preconditioned} camera optimization technique improves camera estimates which results in a higher fidelity scene reconstruction. PSNRs for each patch are inset.
    }
    \label{fig:teaser}
\end{teaserfigure}

\maketitle
\section{Introduction}

A limiting factor of Neural Radiance Fields (NeRF) is the assumption that camera parameters are known. Typically, Structure-from-Motion (SfM) methods such as COLMAP~\cite{schoenberger2016sfm} are used to recover these parameters, but these techniques can be brittle and often fail when given sparse or wide-baseline views. Even given hundreds of views of the same scene, incremental SfM reconstruction may ignore parts of the scene by filtering feature matches that do not align well with the current model.

Given that Neural Radiance Fields are differentiable image formation models, one can optimize camera parameters alongside the scene structure by backpropagating gradients from the loss to the camera parameters. Earlier camera-optimizing NeRF methods focused on forward-facing scenes with unknown poses~\cite{wang2021nerfmm} but known intrinsics, and initialized the joint optimization by assigning every camera to the identity pose~\cite{lin2021barf}. 

Joint recovery of scene and camera parameters is an ill-conditioned problem which is prone to local minima. This can be partially ameliorated with a coarse-to-fine approach. For MLPs, the higher frequency bands of a positional encoding can be windowed~\cite{park2021nerfies}, e.g., in BARF~\cite{lin2021barf}. For grid-based approaches (e.g., Instant NGP~\cite{muller2022instant}) the influence of higher resolution levels can be decreased~\cite{heo2023robust,melas2023realfusion}. 
While these coarse-to-fine approaches are effective, they do not eliminate the need for roughly accurate camera poses for initialization.

An often overlooked aspect of camera-optimizing NeRFs is the choice of the camera parameterization. For example, BARF~\cite{lin2021barf} represents pose as an $\sethree$ 6D screw-axis parameterization, while SCNeRF~\cite{jeong2021self} uses a Gram-Schmidt based 6D representation of orientation~\cite{zhou2019continuity}, plus a 3D translation. The choice of which coordinate frame to represent the translation also matters, and works in 3D object pose estimation have studied egocentric (relative to the camera) and allocentric representations (relative to the world) ~\cite{kundu20183drcnn, li2018deepim}, as well as the interaction between focal length and translation~\cite{ponimatkin2022focalpose}. Moreover, obtaining good estimates of the camera intrinsics (focal length, principal point and distortion coefficients) is key to accurate reconstruction, and SfM systems typically estimate them including radial and tangential distortion coefficients. Intrinsic parameters can be optimized using the same backpropagation technique, as done in SCNeRF.

Given the large space of possible camera parameterizations, what makes one parameterization better suited than another for joint scene and camera estimation? Analyzing this problem is complex, as the scene reconstruction affects the estimation of camera parameters and vice-versa, and the optimization dynamics of neural scene representations (\eg MLPs in NeRF) are challenging to model. We instead analyze the effects of the camera parameterization on a simpler \emph{proxy} setting: how does the camera parameterization affect the projection of points in front of it? Despite its simplicity, this proxy problem still yields insights; it demonstrates that a naively implemented camera parameterization can be ill-conditioned, where perturbations on one axis of the camera parameterization affect the projection significantly more than others.

Although some of these conditioning issues can be addressed on a piecemeal basis (for example, by manually introducing a scaling hyperparameter for each dimension of the camera parameterization) this can still leave spurious correlations between parameters that still contribute to the poorly-conditioned nature of the problem. We draw inspiration from numerical methods, and design a \emph{preconditioner} for each camera in the optimization problem that normalizes the effects of each parameter with respect to the projection of points in the scene, and decorrelates each parameter's effects from all others. This preconditioner is a $k\times k$ matrix (where $k$ is the number of camera parameters) that is applied on the camera parameters before passing them to the NeRF model.
We call this technique \modelname (\nameacronym).

 We base our experiments on the state-of-the-art method of Zip-NeRF~\cite{barron2023zip}, that uses an anti-aliasing sampling strategy with multi-resolution hash grids~\cite{muller2022instant}. We evaluate \nameacronym on real captures from the mip-NeRF 360 dataset~\cite{barron2022mipnerf360}, and as well on perturbed versions of the the mip-NeRF 360 and the synthetic NeRF scenes~\cite{mildenhall2020nerf}. We show that our preconditioning strategy is effective when using different camera parameterizations, including the one of~\citet{jeong2021self}, and FocalPose~\cite{ponimatkin2022focalpose}. Finally, we show that \nameacronym also improves the reconstruction of challenging cellphone sequences with ARKit poses.

\begin{figure*}
    \centering
	\captionsetup[sub]{labelformat=parens}
	\begin{subfigure}[b]{0.27\textwidth}
        \centering
    	\includegraphics[height=43mm, clip, trim=50 620 1420 0]{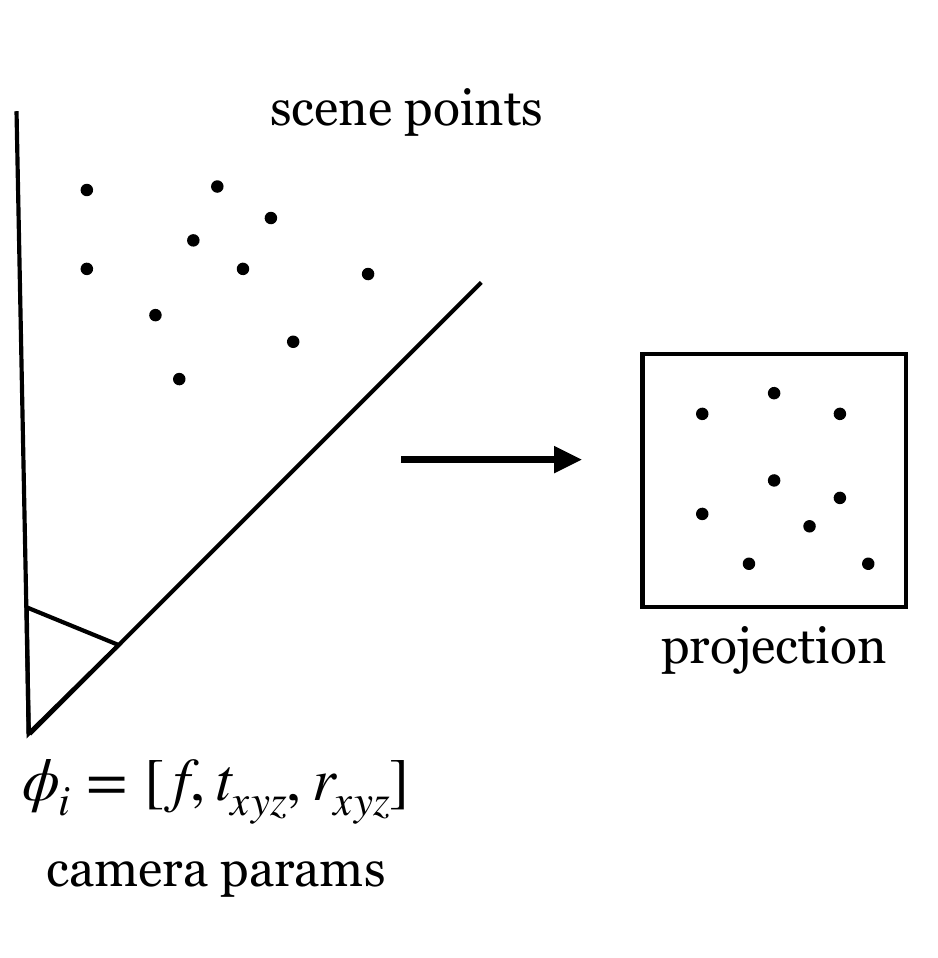}
    	\subcaption{Proxy Projection}
        \label{fig:preconditioning_proxy_projection}
	\end{subfigure} 
	\begin{subfigure}[b]{0.36\textwidth}
        \centering
    	\includegraphics[height=43mm, clip, trim=540 620 770 30]{figures/overview_figure_v1.pdf}
    	\subcaption{Camera Parameter Decorrelation (\namehandle)}
        \label{fig:preconditioning_decorrelation}
	\end{subfigure}
	\begin{subfigure}[b]{0.36\textwidth}
        \centering
    	\includegraphics[height=43mm, clip, trim=1210 620 180 30]{figures/overview_figure_v1.pdf}
    	\subcaption{Preconditioned NeRF}
        \label{fig:preconditioning_nerf}
	\end{subfigure}
    \caption{Overview of our method. (a) We use the projection of randomly sampled scene points within the frustum of camera $i$ as a proxy problem (b) to compute the covariance matrix $\mat{\Sigma}_i$ of the derivatives of the projection of those points with respect the individual camera parameters, and with that we derive a per-camera preconditioner $\mat{P}_i^{-1}$. (c) This lets us reformulate the naive optimization problem (wherein optimization is performed directly on camera parameters $\{\mat{\phi}_i\}$) into one in which we instead optimize over a set of latent preconditioned parameters $\{\Tilde{\mat{\phi}}_i\}$: each time we compute the loss function we multiply each $\Tilde{\mat{\phi}}_i$ by its corresponding $\mat{P}_i^{-1}$ matrix to yield camera parameters $\mat{\phi}_i$ for use as input to the NeRF.}
    \label{fig:overview}
\end{figure*}
\section{Related Work}

\paragraph{3D Reconstruction}
Recovering a scene from images is a long-standing problem in computer vision. Structure-from-Motion (SfM) is a central technique, where sparse keypoint matches are triangulated and bundle adjusted to minimize reprojection metrics~\cite{schoenberger2016sfm}.
Most recently, learning-based techniques have been used to recover more accurate representations, such as methods based on rendered proxies and real-time convolutional neural networks that produce the final image~\cite{meshry2019neural, thies2019deferred}, or blending weights~\cite{hedman2018deep}. Neural Radiance Fields (NeRF)~\cite{mildenhall2020nerf} encode a scene in the weights of an MLP and use volume rendering to produce highly realistic novel views of a captured scene. NeRFs have been extended to a variety of tasks such as internet photo collections~\cite{martinbrualla2020nerfw}, dynamic scenes~\cite{li2020nsff, park2021nerfies}, and lighting estimation~\cite{bi2020neural}.

\paragraph{Camera-optimizing NeRFs}
Differentiable rendering methods, such as the volume rendering used in NeRF, enable backpropagation through the scene representation to update the camera parameters. While these have been explored in the context of meshes and rasterization~\cite{loper2014opendr, liu2019soft}, we focus on approaches applicable to NeRF-like models. NeRF$-$ \!$-$~\cite{wang2021nerfmm} shows that poses can be estimated for forward facing scenes by initializing cameras to the origin. BARF~\cite{lin2021barf} uses a coarse-to-fine reconstruction scheme~\cite{park2021nerfies, hertz2021sape} that gradually introduces higher-frequency of position encodings. Coarse-to-fine approaches can be adapted to multi-resolution grids such as NGP~\cite{muller2022instant} by using a weighing schedule of the different resolution levels~\cite{melas2023realfusion, heo2023robust}. GNeRF \cite{meng2021gnerf} proposes a pose-conditioned GAN that is then used to recover a NeRF. Others have explored more suitable architectures to perform pose optimization, such as Gaussian ~\cite{chng2022garf} or sinusoidal activations~\cite{xia2022sinerf}.
NoPe-NeRF~\cite{bian2022nope} uses monocular depth priors to constrain the scene as well as relative pose estimates. Keypoint matches or dense correspondences can also be used to constrain the relative pose estimates using ray-to-ray correspondence losses~\cite{truong2022sparf, jeong2021self}.
DBARF~\cite{chen2023dbarf} proposes using low-frequency feature maps to guide the bundle adjustment for generalizable NeRFs~\cite{wang2021ibrnet, yu2021pixelnerf}.

Although many methods ignore the optimization of intrinsic parameters, these are key to achieving high quality reconstructions. SCNeRF~\cite{jeong2021self} models a residual projection matrix, and residual raxel parameters~\cite{grossberg2005raxel}, which are interpolated on a sub-sampled pixel grid~\cite{zhou2014color}.

NeRFs have also been adapted to real-time SLAM settings by using NeRFs to accumulate scene observations in an online training fashion~\cite{rosinol2022nerfslam, zhu2022nice}.
LocalRF~\cite{meuleman2023progressively} reconstructs long trajectories without relying on SfM by using an incremental strategy that adds images to the reconstruction one at a time, and a subdivision strategy similar to BlockNeRF~\cite{tancik2022block}. Notably LocalRF uses different learning rates for translation and orientation parameters, highlighting the challenges in taming camera-optimizing NeRFs.

\paragraph{Preconditioning}
Many computer graphics tasks rely on solving large-scale linear systems or optimizing non-linear objectives, such as the bundle adjustment in SfM. Preconditioned methods for linear systems enable faster convergence, but finding the best preconditioner for a task is hard, as one must weigh the advantages of the preconditioner against the time required to compute it. Notably, preconditioners can be designed using domain knowledge of a given task, such as the visibility-structure of a matching graph in SfM~\cite{kushal2012visibility}. We refer the reader to~\citet[A.5.2]{szeliski2022computer} for a review.

In the case of stochastic gradient descent (SGD), preconditioning is equivalent to gradient scaling~\cite{DeSa2019LectureNotes}.
Gradient scaling has been recently proposed in the context of NeRF optimization, to reduce floaters close to the cameras, as those areas are oversampled relative to the background~\cite{philip2023radiance}.

\paragraph{Camera Parameterizations}
The choice of camera parameterization for representing the image formation model can have significant effects in applications that predict or optimize over those parameters. For instance, some parameterizations of rotation are well known to not be continuous or differentiable over the $\SOTHREE$ manifold (\eg, Euler angles). 
In addition, 3D reconsruction tasks suffer from ambiguities, such as the gauge ambiguity where the error is invariant up to a similarity transform, ambiguities between depth and focal length for planar scenes, and complex ambiguities between radial distortion and camera position~\cite{wu2014critical}. These ambiguities can lead to instability during optimization.
Finally, many image formation models, such as traditional lens distortion models, are low-order approximations to more complex phenomena, and can significantly under-fit the signal assuming sufficient data is available~\cite{schops2020having}.

Several works have been proposed to address these issues.
\citet{zhou2019continuity} propose a 6D continuous parameterization of rotations based on Gram-Schmidt orthogononalization. Early works on 3D pose estimation~\cite{kundu20183drcnn, li2018deepim} find that highly correlated rotation and translation parameters lead to poor results, and propose object-centric rotation and translation representations that decorrelate them. This idea is further extended to joint pose and focal length estimation by~\citet{ponimatkin2022focalpose}.
In our work, we first analyze the structure of the parameter space of camera representations by measuring their effects on the projections of 3D points in the scene, and then propose a preconditioning scheme for camera optimization to normalize and decorrelate the parameters during optimization.

\fboxsep=0pt %
\fboxrule=0.4pt %

\newcommand{\blurimsz}{0.138}
\begin{figure*}
    \centering
	\begin{subfigure}[b]{0.87\textwidth}
        \centering
    	\figcellt{\blurimsz}{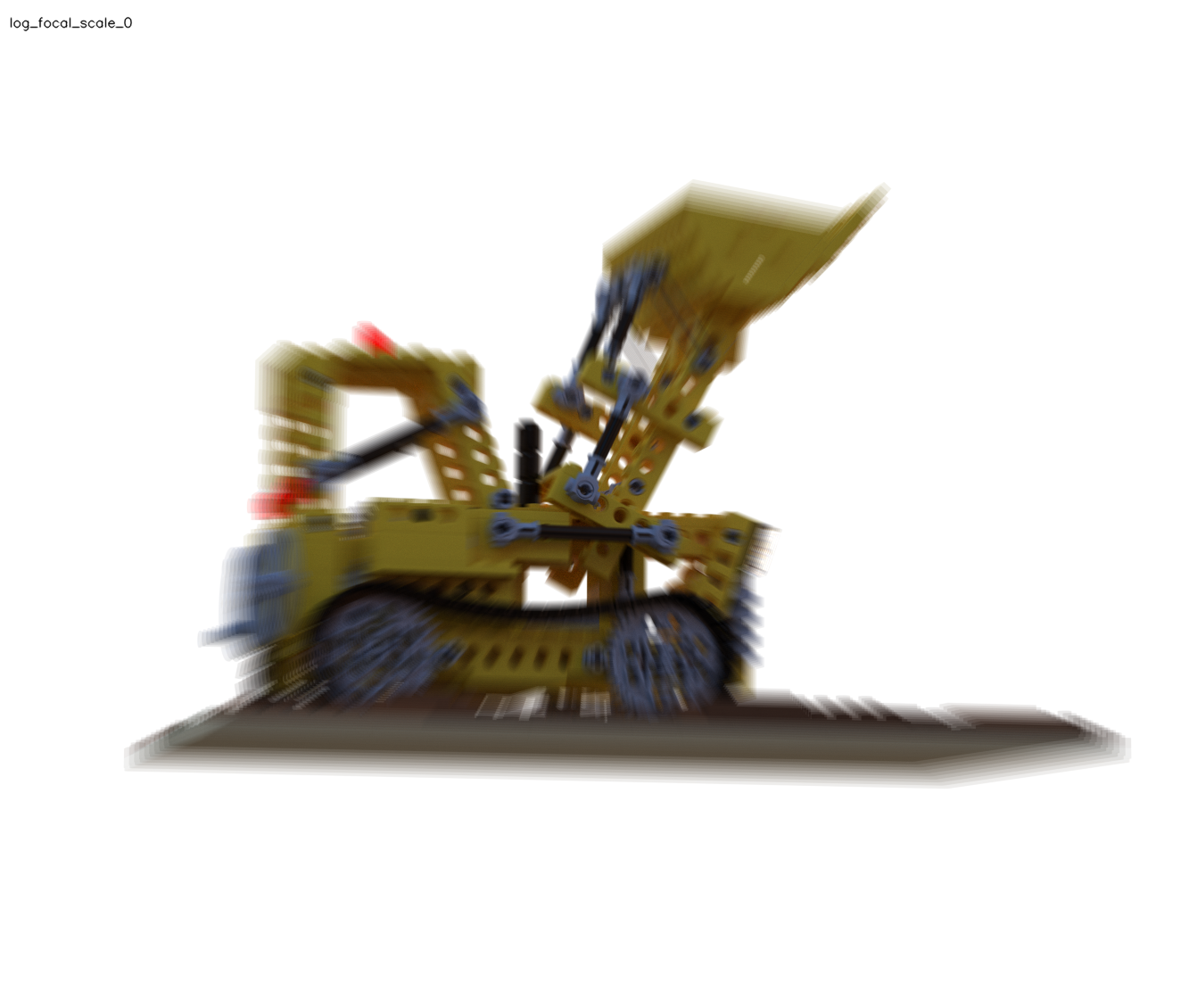}{}{clip,trim=166 65 166 65}
    	\figcellt{\blurimsz}{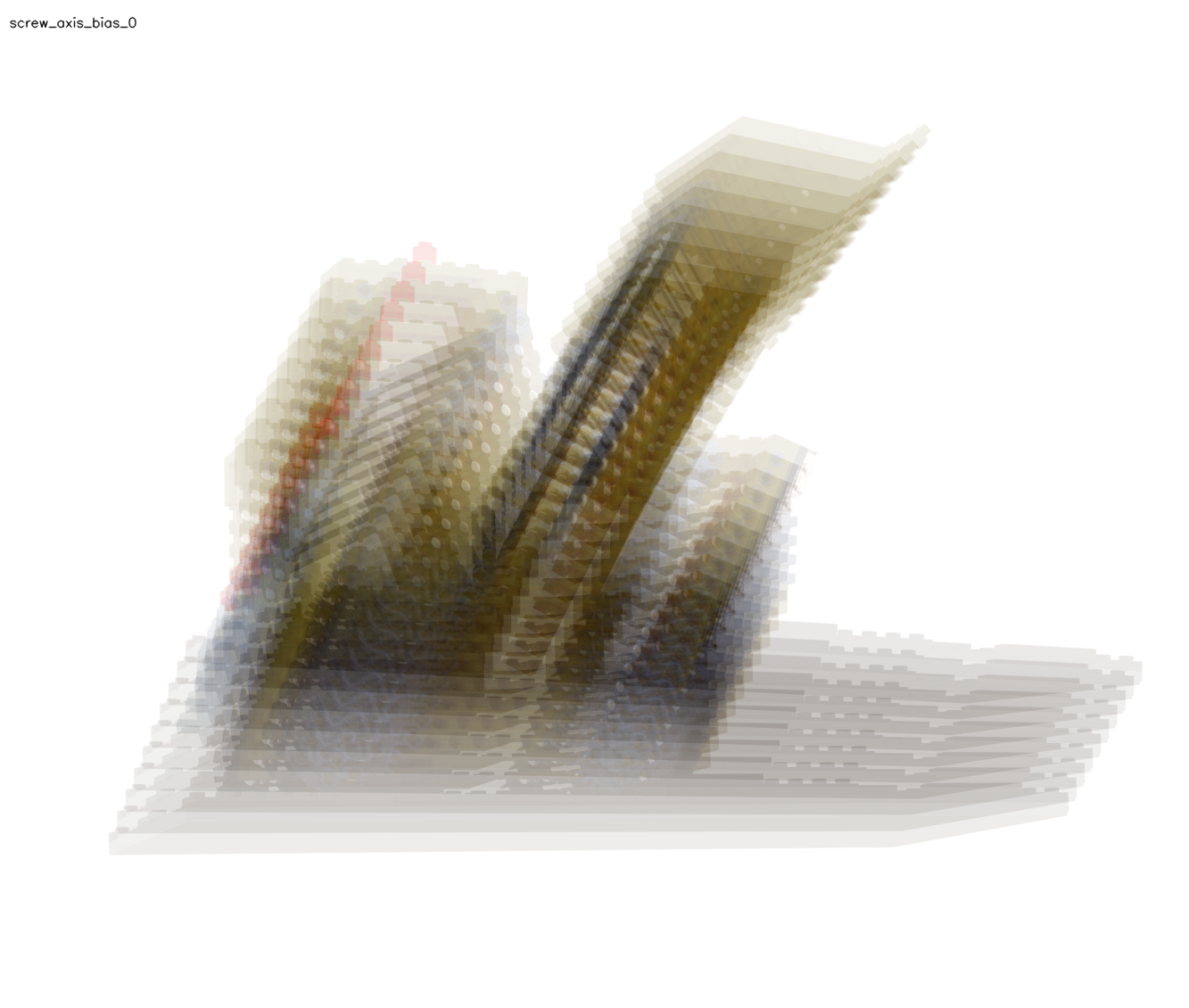}{}{clip,trim=166 65 166 65}
    	\figcellt{\blurimsz}{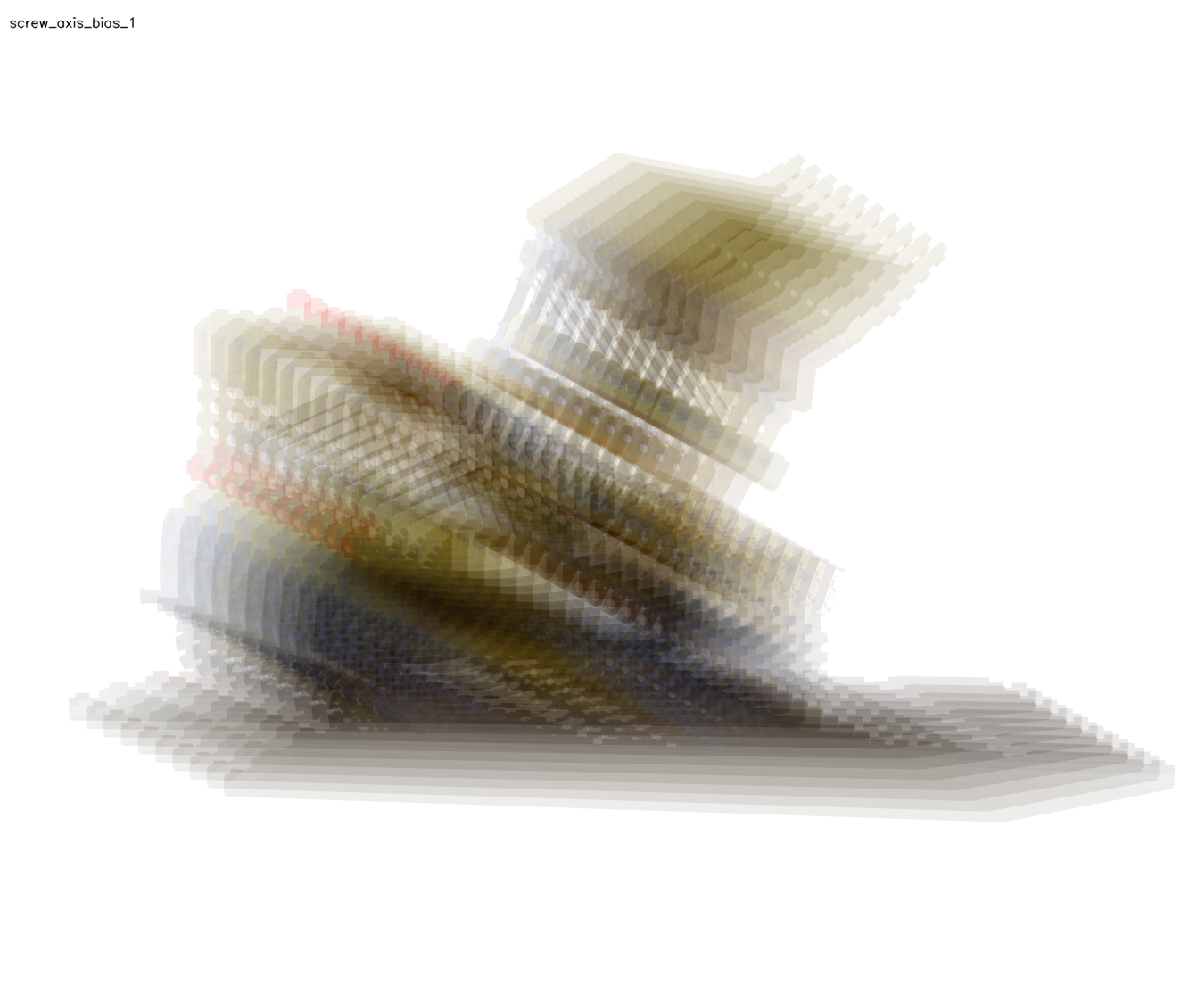}{}{clip,trim=166 65 166 65}
    	\figcellt{\blurimsz}{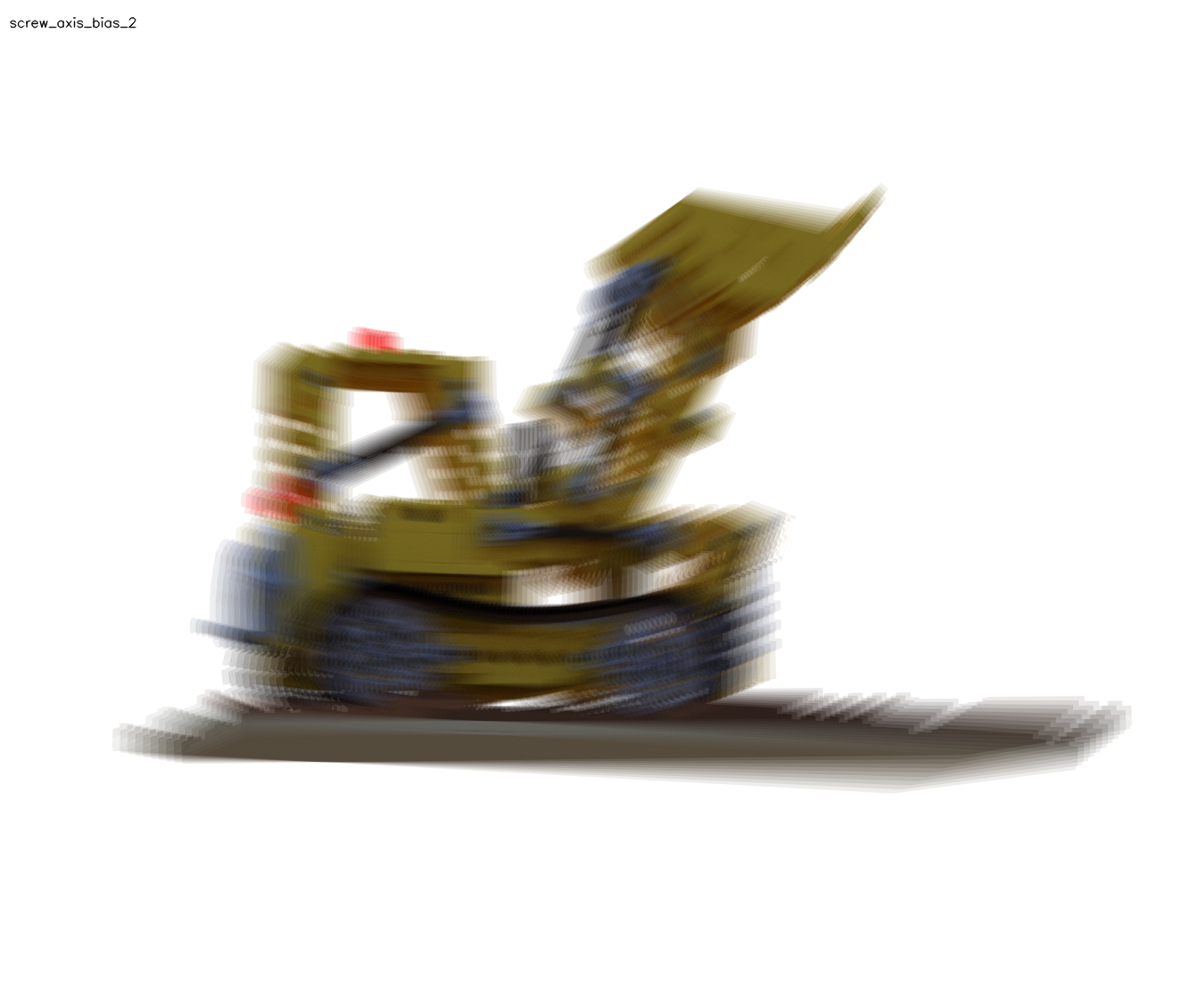}{}{clip,trim=166 65 166 65}
    	\figcellt{\blurimsz}{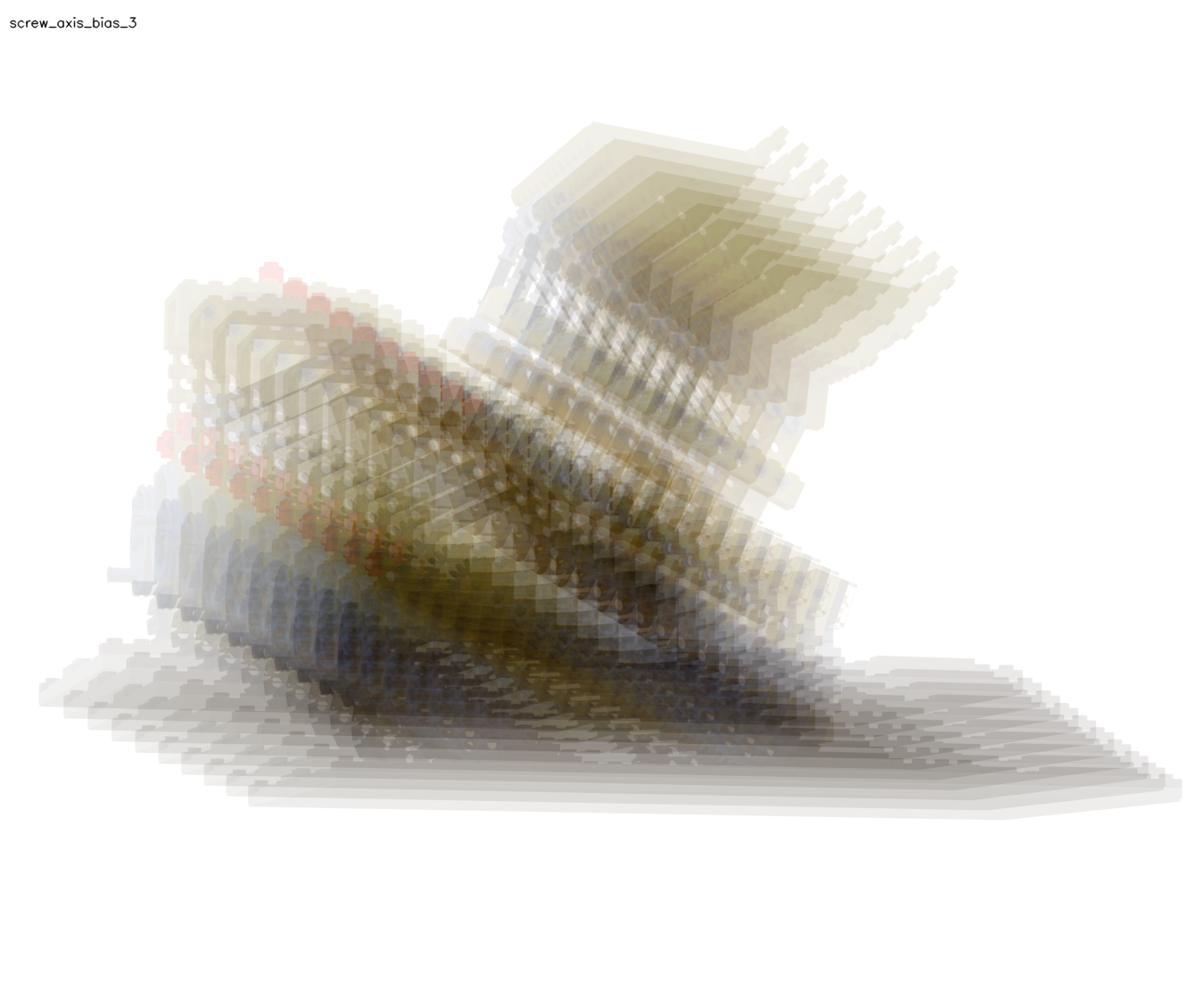}{}{clip,trim=166 65 166 65}
    	\figcellt{\blurimsz}{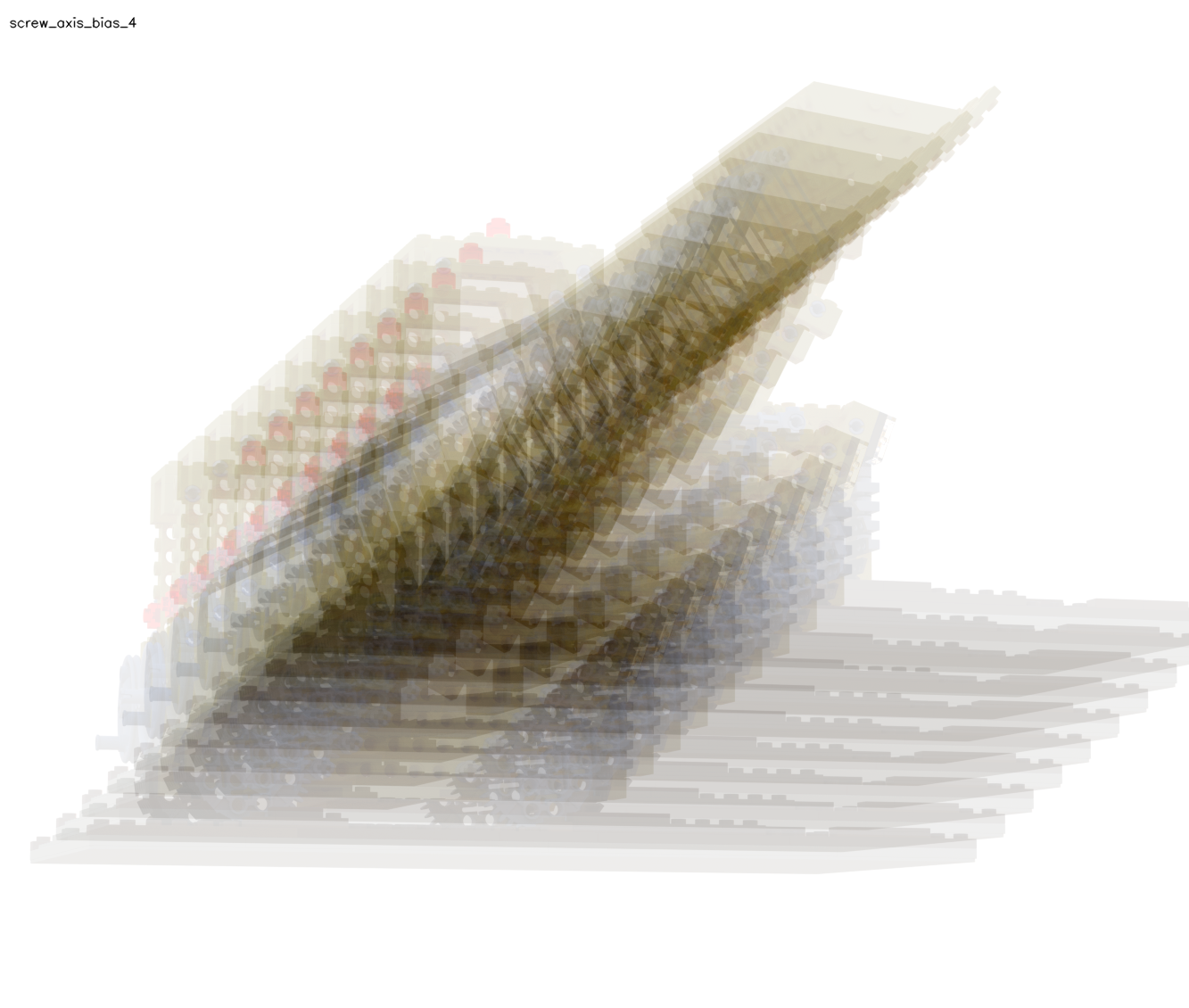}{}{clip,trim=166 65 166 65}
    	\figcellt{\blurimsz}{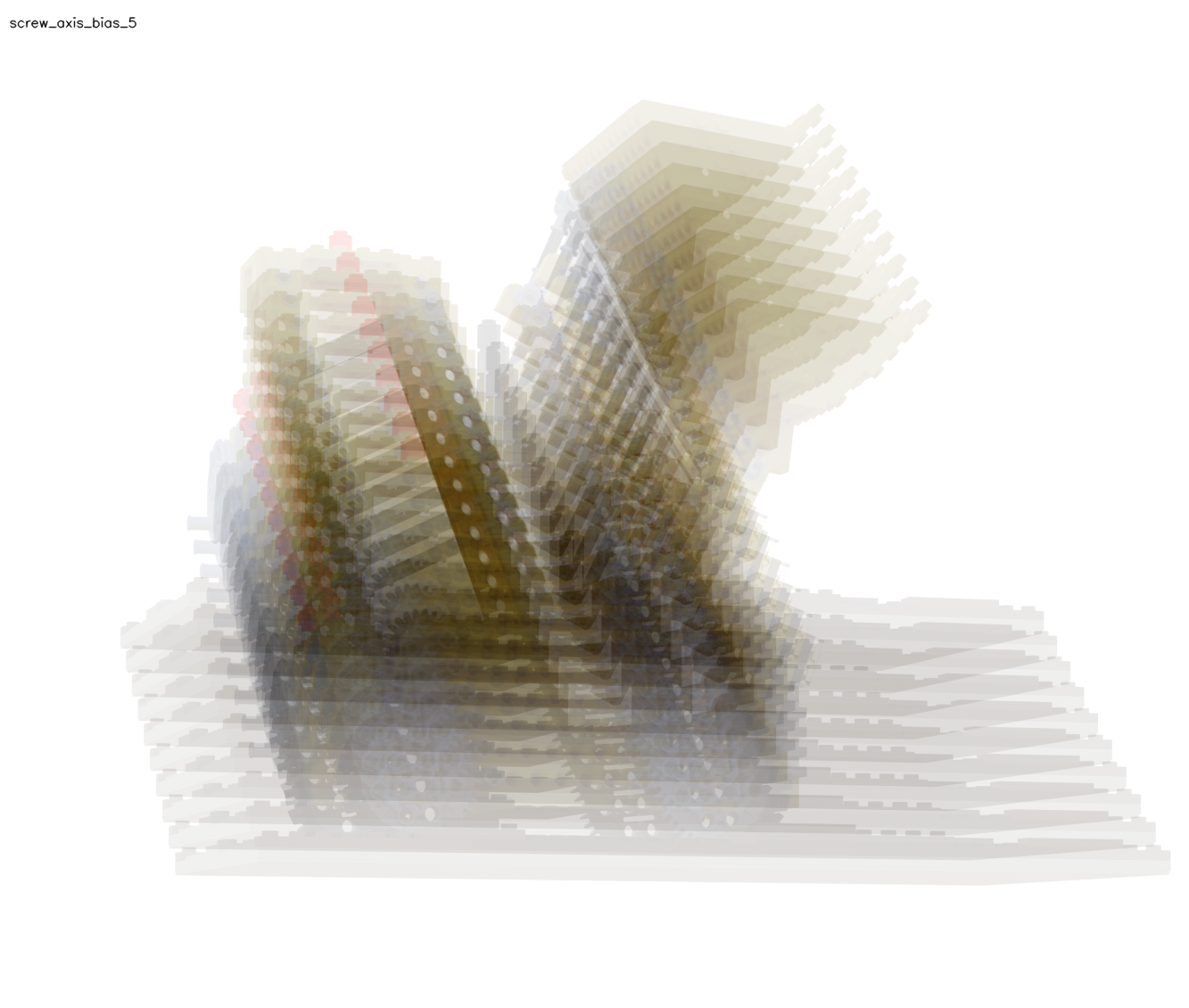}{}{clip,trim=166 65 166 65}
    	\subcaption{Motion trails for \camerafont{SE3+Focal}}
        \label{fig:se3focal_motion_trails}
	\end{subfigure}%
	\begin{subfigure}[b]{0.13\textwidth}
        \centering
    	\figcellt{0.925}{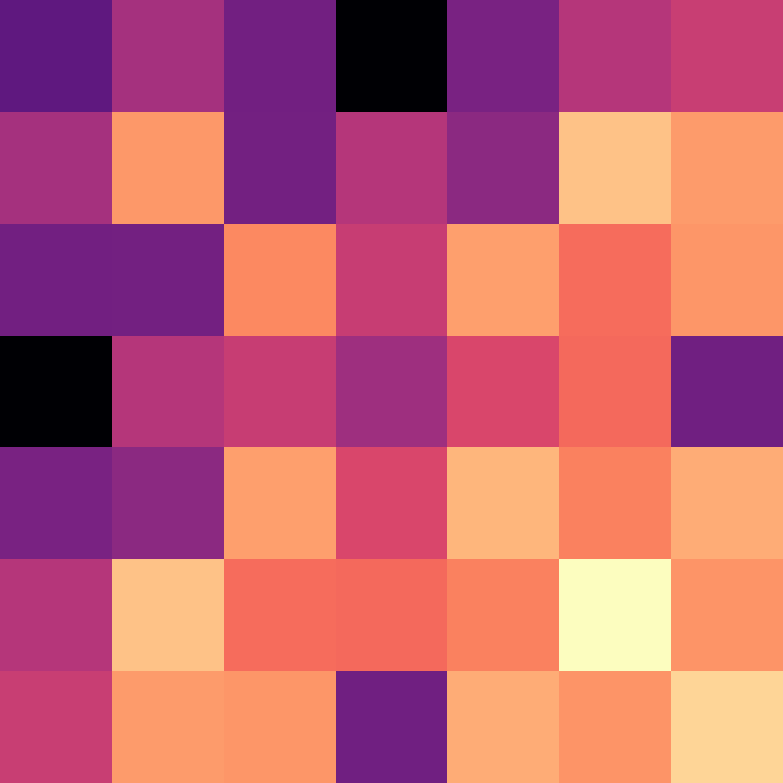}{}{}
    	\subcaption{Covariance}
        \label{fig:covariance}
	\end{subfigure}

    \vspace{1mm}
	\begin{subfigure}[b]{0.87\textwidth}
        \centering
    	\figcellt{\blurimsz}{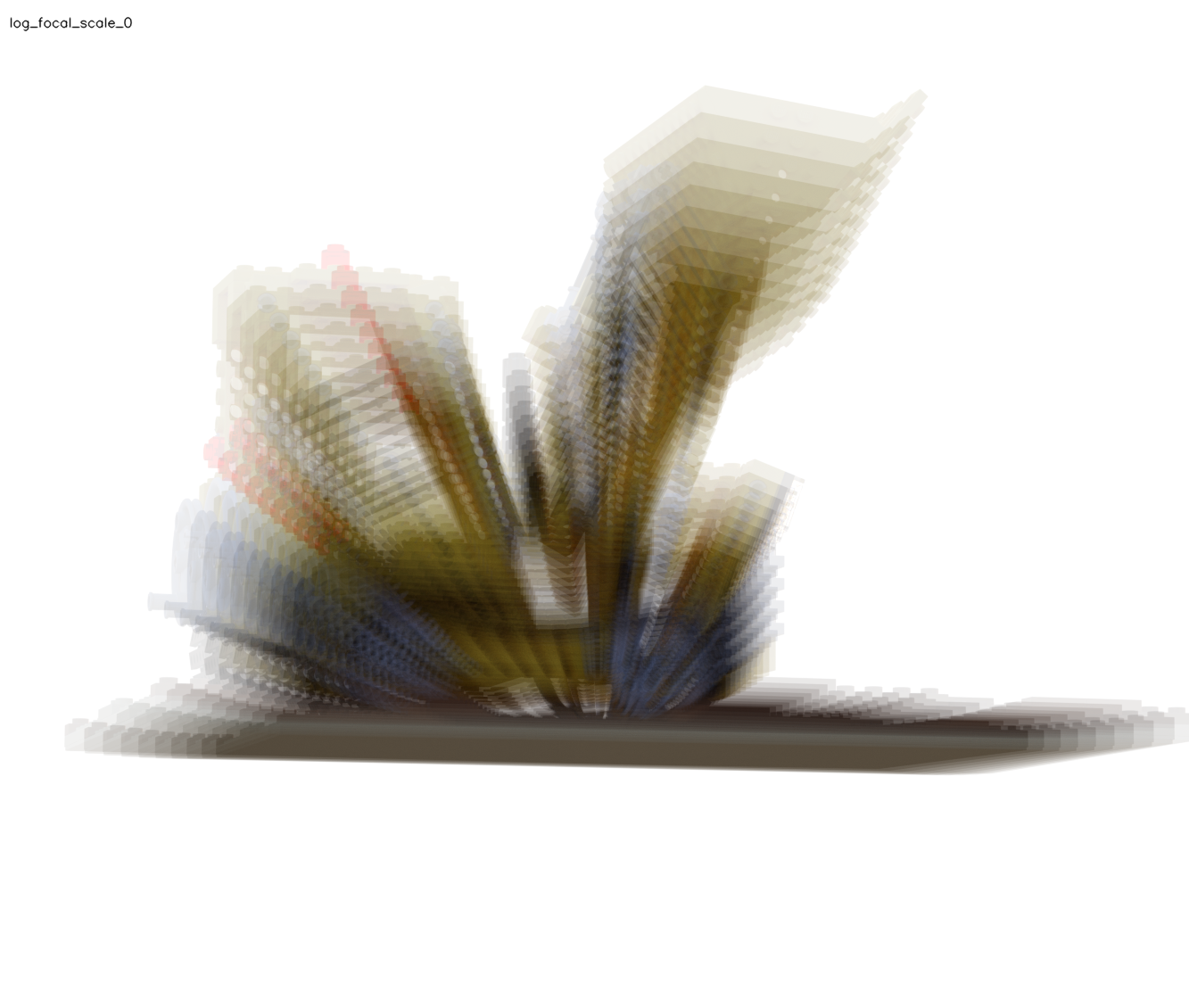}{focal}{clip,trim=166 65 166 65}
    	\figcellt{\blurimsz}{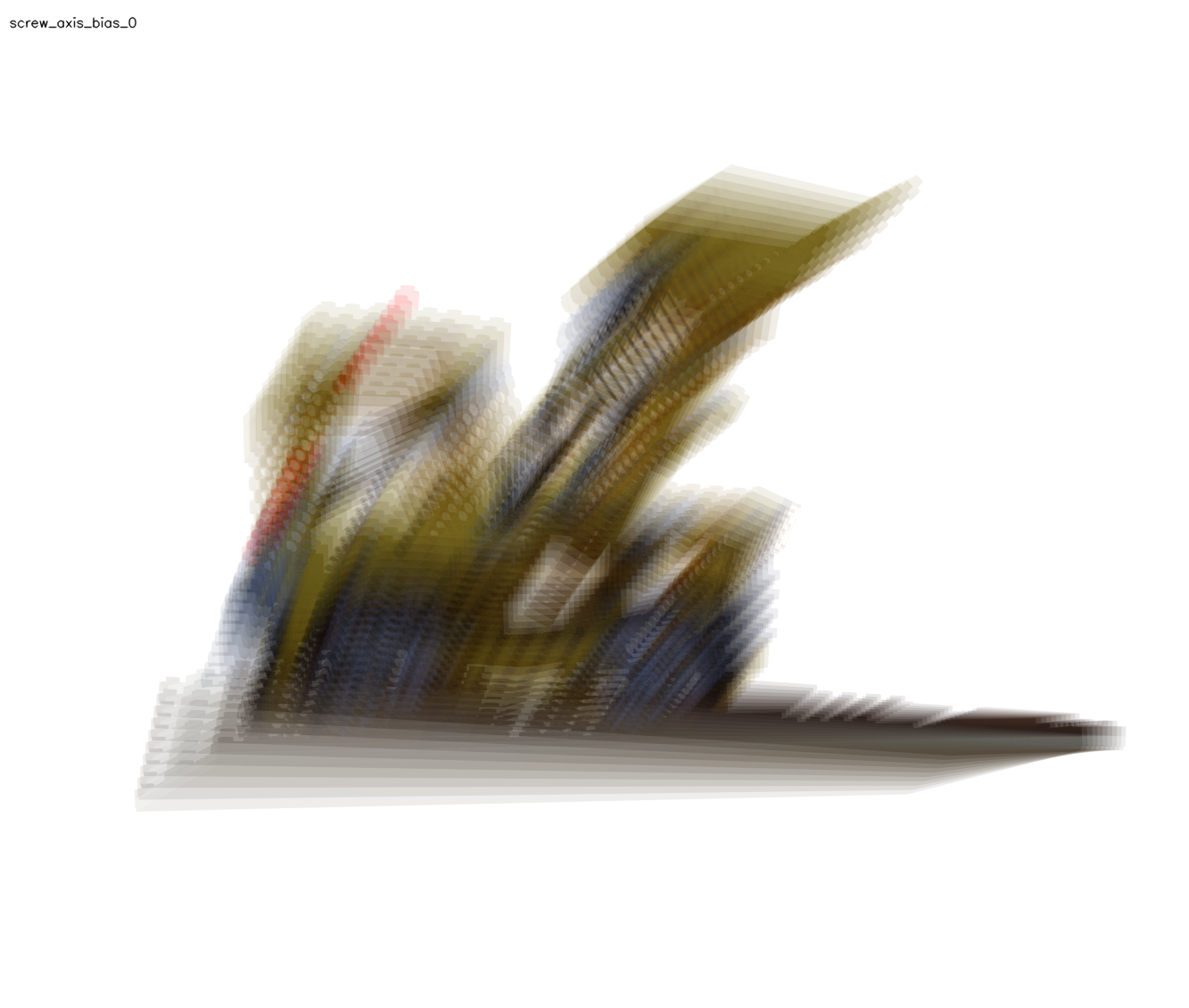}{$x$ rotation}{clip,trim=166 65 166 65}
    	\figcellt{\blurimsz}{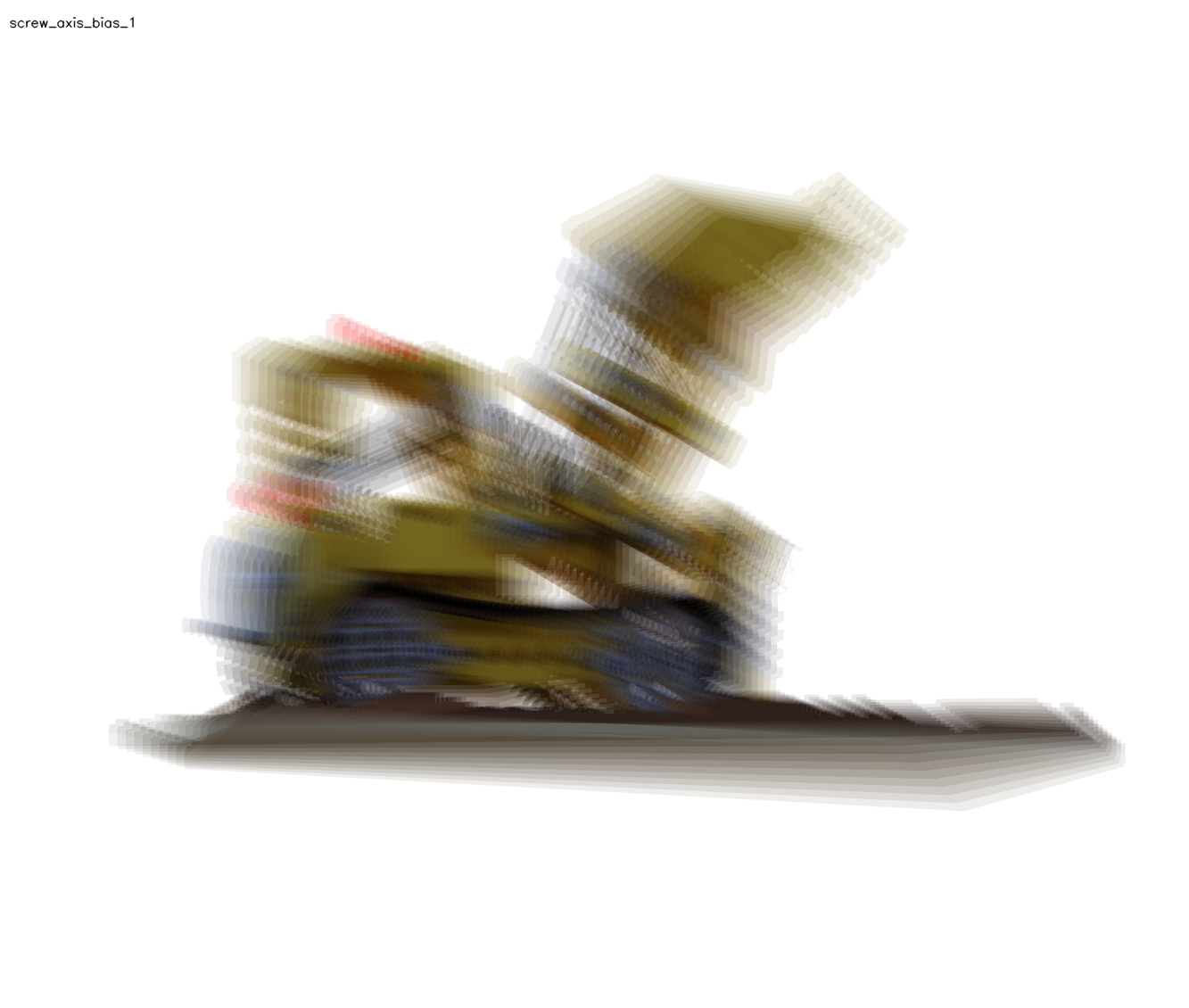}{$y$ rotation}{clip,trim=166 65 166 65}
    	\figcellt{\blurimsz}{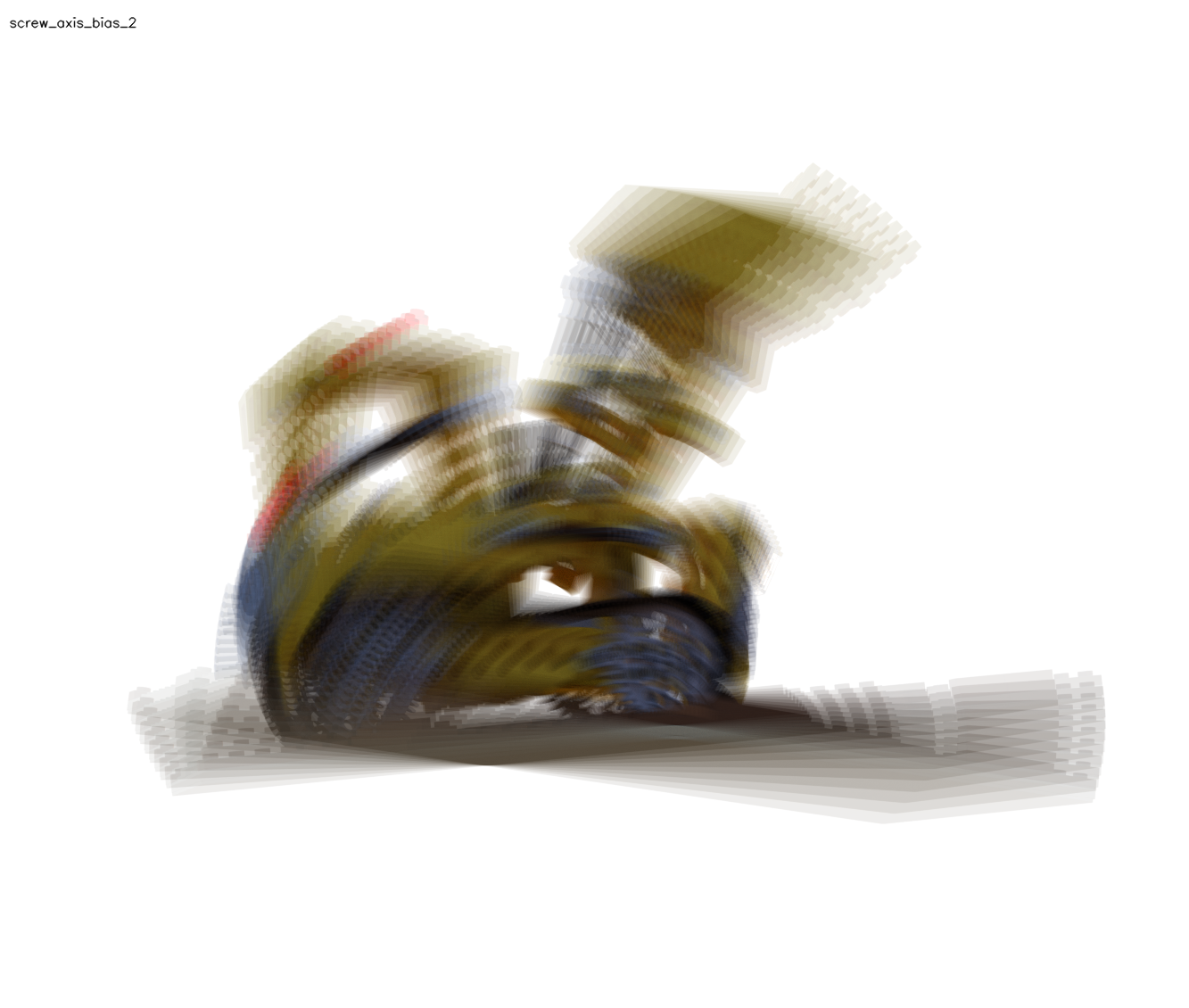}{$z$ rotation}{clip,trim=166 65 166 65}
    	\figcellt{\blurimsz}{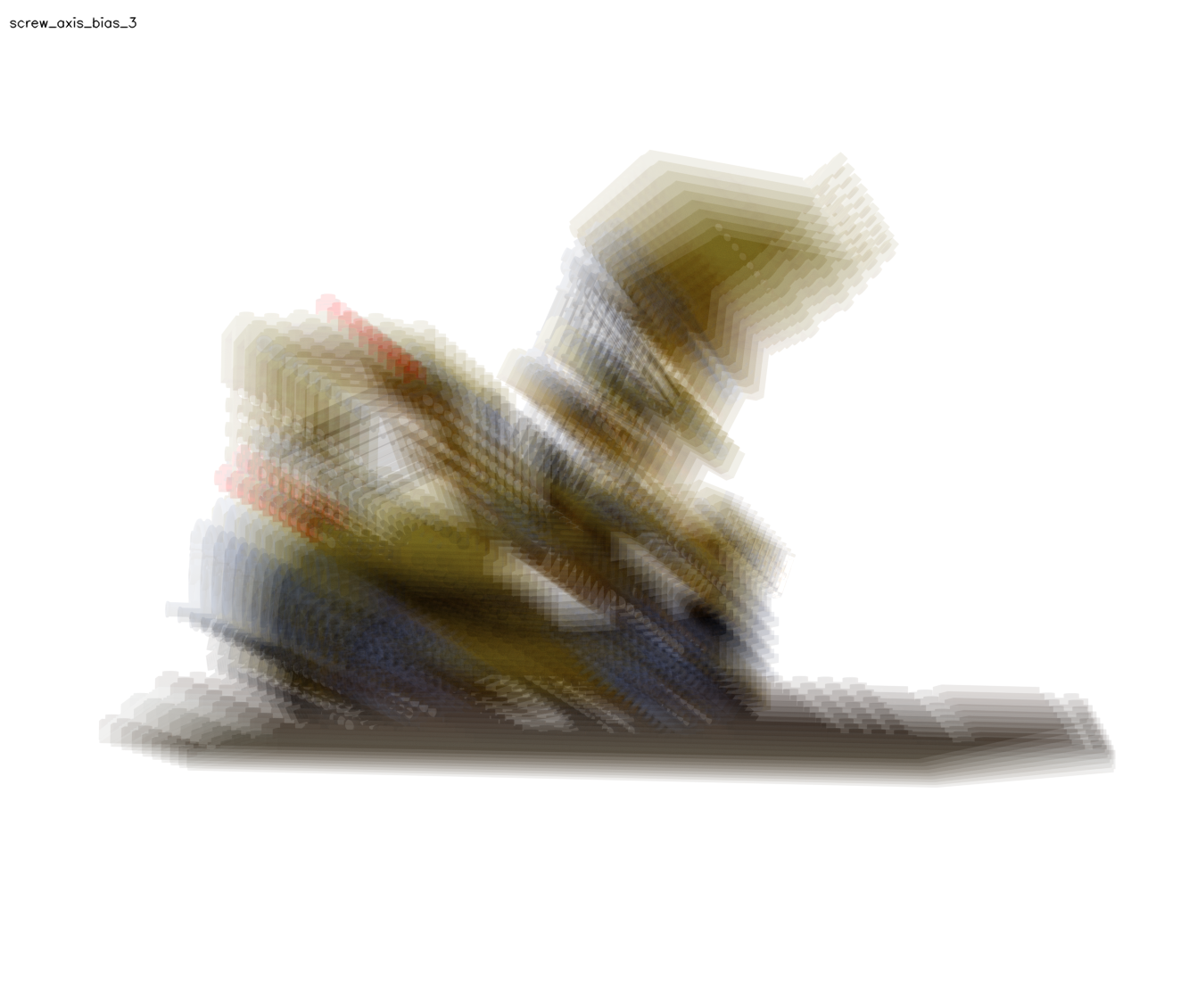}{$x$ translation}{clip,trim=166 65 166 65}
    	\figcellt{\blurimsz}{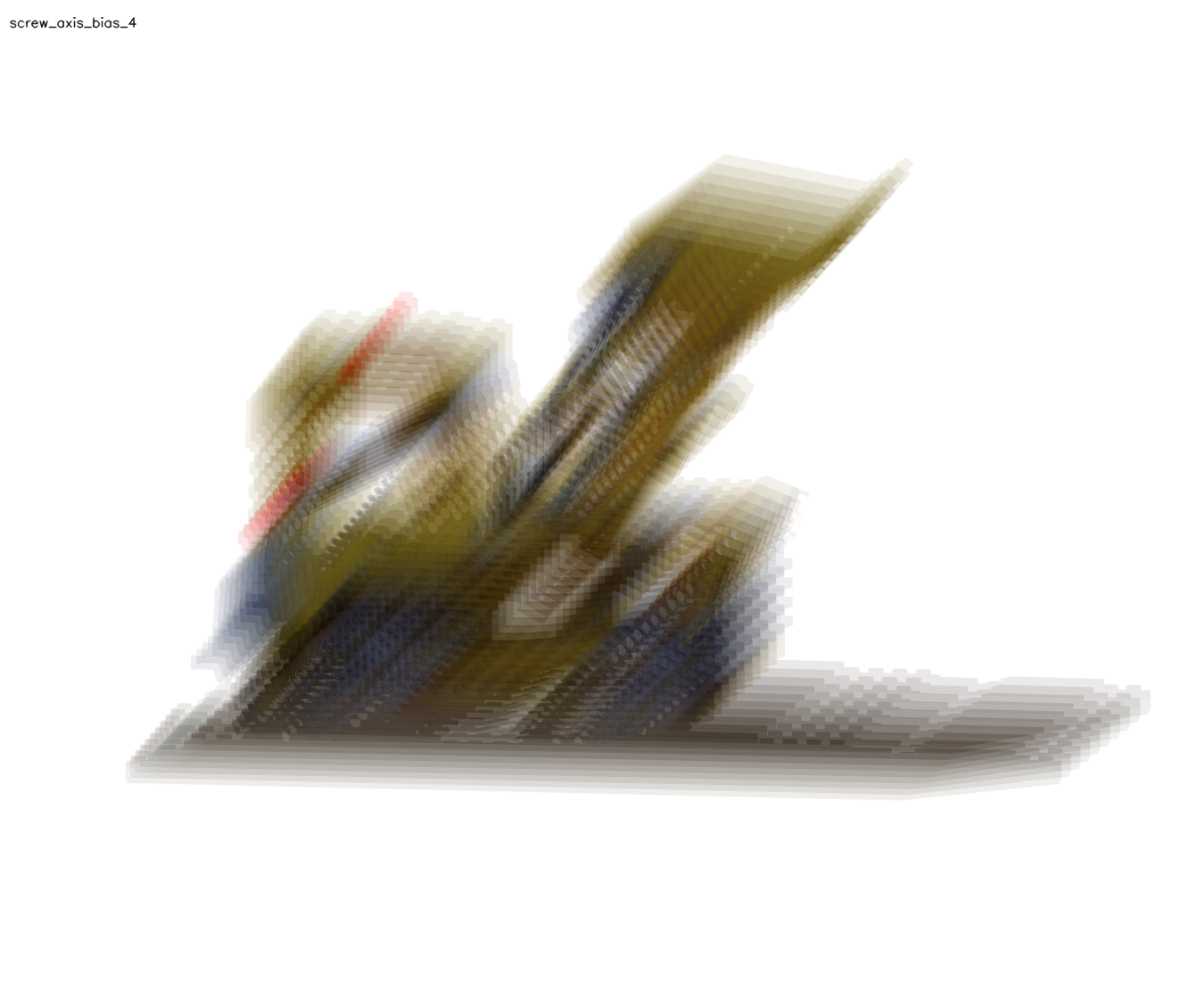}{$y$ translation}{clip,trim=166 65 166 65}
    	\figcellt{\blurimsz}{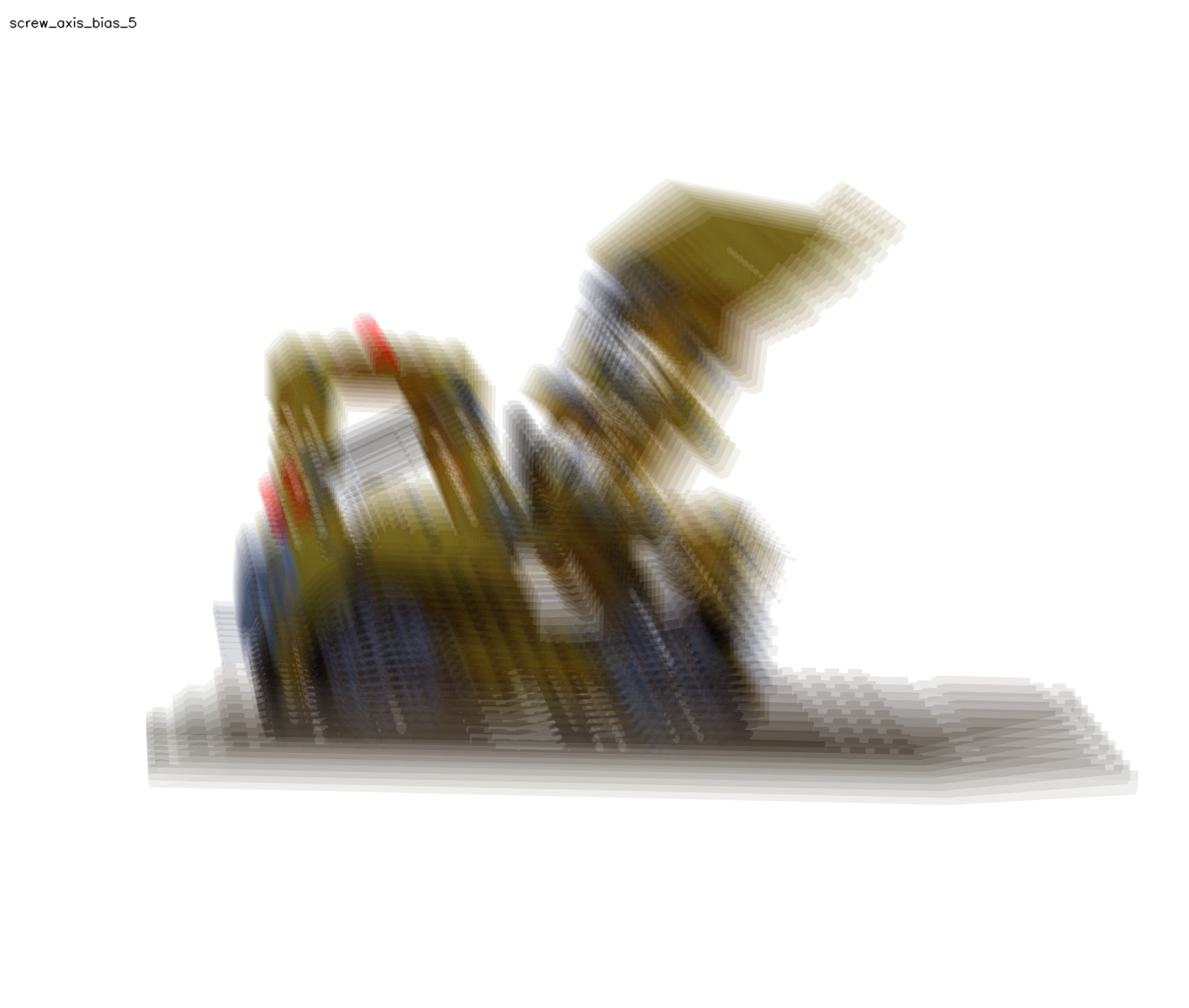}{$z$ translation}{clip,trim=166 65 166 65}
    	\subcaption{Motion trails for \camerafont{SE3+Focal} with \textbf{\nameacronym}}
        \label{fig:preconditioned_se3focal_motion_trails}
	\end{subfigure}%
	\begin{subfigure}[b]{0.13\textwidth}
        \centering
    	\figcellt{0.925}{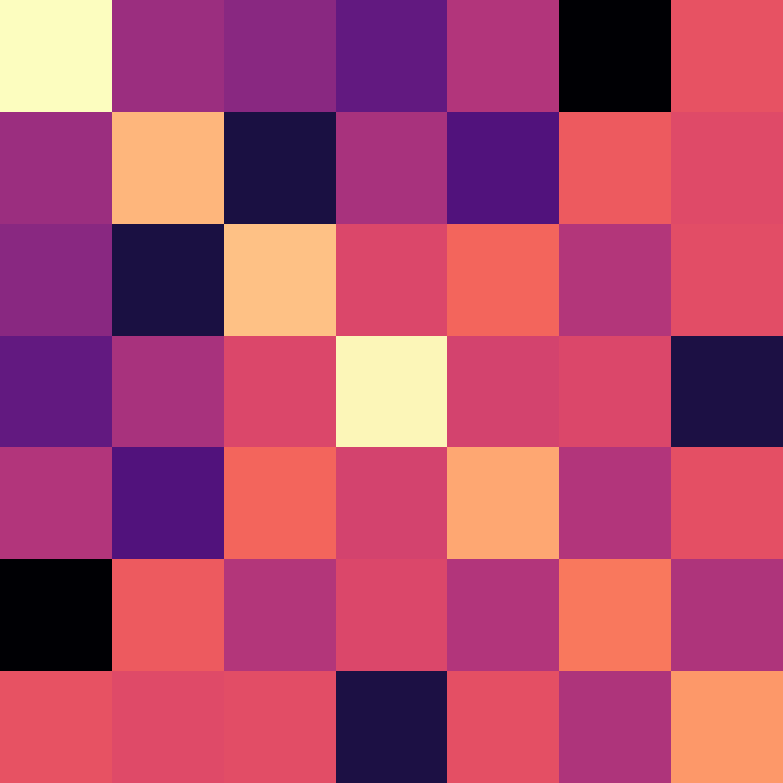}{~}{}
    	\subcaption{Preconditioner}
        \label{fig:preconditioner}
	\end{subfigure}

    \caption{We analyze the effect of perturbing individual camera parameters for our \camerafont{SE3+Focal} camera parameterization, by rendering the motion trails produced by varying focal length, rotation, and translation. (a) Motion magnitudes differs significantly when varying these camera parameters: \eg, changing focal length causes little motion, while changing translation results in large motions. (c) Preconditioning the camera parameters results in motion magnitudes that are similar across all parameters. We also visualize the (b) covariance of the proxy-projection and (d) preconditioning matrices constructed by our model.}
    \label{fig:preconditioning}
    \vspace{-1em}
\end{figure*}

\section{Preliminaries}
\subsection{Camera-optimizing NeRFs}

We are given a set of $n$ images $\left\{I_i\right\}$ and initial camera parameters $\left\{\varcamparam_i^0\right\}$, where $i\in \{ 1, \ldots, n \}$.
Our goal is to obtain a scene model $\varmodel_{\varmodelparam}$ with optimized NeRF parameters $\varmodelparam^*$ and optimized camera parameters $\varcamparam_i^*$, that best reproduce the observed images.
The NeRF is trained by minimizing a loss function $\varloss(D; \varmodelparam, \varcamparam)$ over a dataset $D$ containing all observed pixels, where pixel $j$ has a camera index $d_j \in \{ 1, \ldots, n \}$, a pixel location $\varpixel_j$, and an RGB color $\varpixelvalue_j$.

Similar to non-linear least square methods, we optimize over a linearized version of the problem for the camera parameters $\varcamparam = \varcamparam^0 + \vardeltacamparam$, where we optimize over a residual of the camera parameters $\vardeltacamparam$ w.r.t. their initial values $\varcamparam^0$. We denote the dimensionality of the camera parameterization as $k$, i.e., $\varcamparam_i \in \varR^k$

A camera parameterization defines a projection function $\Pi$ where $\Pi(\varpoint; \varcamparam): \varRthree \rightarrow \varRtwo$ projects a 3D point $\varpoint$ to a 2D pixel location $\varpixel$, with a corresponding inverse that maps pixels to points $\Pi^{-1}(\varpixel, d; \varcamparam): \varRtwo \times \varR \rightarrow \varRthree$, where $d$ is the pixel depth.

\subsection{Preconditioning}
\newcommand{\fakedim}{k}
Preconditioning\footnote{This is a slight abuse of terminology since the strictest definition of preconditioning only applies to linear systems with a condition number.}~\cite[A.5.2]{szeliski2022computer} is a technique where an optimization problem is transformed using a \emph{preconditioner} to have better optimization characteristics, \eg, one with a lower condition number.
For instance, when optimizing a value $\mat{u} \in \varR^\fakedim$ to minimize $f(\mat{u}): \varR^n \rightarrow \varR$, we can precondition the system by computing $\mat{v} = \mat{P}\mat{u}$ (where $\mat{P}\in\varR^{\fakedim\times \fakedim}$ is the preconditioning matrix) and instead optimizing $f(\varprecon\mat{v})$. Once an optimal $\mat{v}^*$ is found, one can obtain $\mat{u}$'s optimal value  as $\mat{u}^*= \varprecon\mat{v}^*$.
Ideally, $\varprecon$ approximates the inverse Hessian, resulting in the Newton-Raphson method.

In the context of the gradient-based optimizers typically used in NeRF, we can precondition a variable $\mat{a}$ by optimizing over a preconditioned version of the variables $\tilde{\mat{a}} = \mat{P} \mat{a}$, and modifying the objective to be a function of $\varprecon\tilde{\mat{a}}$ instead of $\mat{a}$. If SGD is the optimizer, gradient scaling techniques (such as that of~\citet{philip2023radiance}) are equivalent to preconditioning.

\section{Analyzing Camera Parameterizations}
The choice of camera parameterization has a significant impact on the output reconstruction quality. SfM pipelines such as COLMAP support perspective or fisheye cameras and different lens distortion models, including models with radial and distortion parameters. Modeling these effects is critical when using real-world images.

There are two ways one can think about a camera's extrinsic pose: (i) as an \emph{egocentric} parameterization that models the camera's pose with respect to the camera itself, or (ii) as an \emph{allocentric} parameterization~\cite{kundu20183drcnn} that models the camera with respect to the world or object coordinate frame.
The specific choice of rigid 6D pose parameterization presents further complications: one can use quaternions, axis-angle (Rodrigues' formula~\cite{rodrigues1816}), or 6D continuous pose representations~\cite{zhou2019continuity} to represent the orientation, or the $\sethree$ screw axis representation. 

Inspired by \citet{kundu20183drcnn, li2018deepim}, we visualize how perturbing the parameters of a camera model affects the projected image.
Consider a camera that uses an $\sethree$ screw-axis parameterization $\mathcal{S} = (\logrot; \mat{v})\in\mathbb{R}^6$ to encode the rigid pose~\cite{lynch2017modern}, augmented with an additive bias $\Delta f$ for the focal length. \figref{fig:preconditioning}a shows the \emph{motion trails} of a rendered object, where the camera is altered by perturbing each axis of the parameterization by the same magnitude and the resulting frames are averaged together.
Notice how some parameters impact the rendered image more than others, \eg, the image is less sensitive to focal length perturbations compared to translation. The reason is evident when considering that focal length is in pixel units, while translation is in world units.

This difference in sensitivity is problematic when optimizing using a gradient descent method. If the learning rate is large enough to affect focal length, other parameters will overshoot and oscillate. However, if the learning rate is small, focal length will optimize too slowly. Adaptive gradient methods such as Adam~\cite{kingma2014adam} ameliorate this by normalizing the magnitudes of the updates to a similar range, but may ignore differences \emph{between} parameters: the updates to the focal length should be much larger than position, but Adam's parameters updates all have similar magnitudes.

\section{Camera Preconditioning}

A common way to mitigate differences in magnitude in an optimization setting is to re-parameterize the problem. The simplest example of this is multiplying the parameter by a scalar, \eg, the focal length parameterization $f = f^0 + \Delta f$ can be changed to $f = f^0 + s\Delta f$ where $s$ is a large scalar. 
This change is equivalent to preconditioning the optimization problem with diagonal preconditioner $\mat{P}^{-1}$ equal to the identity, except for the entry on the diagonal corresponding to the focal length, with a value of $s$.

While effective, per-parameter scaling cannot decouple highly correlated coefficients. To design a more comprehensive preconditioner for camera optimization, we study the sensitivity of the camera projection function $\varprojection(\varpoint;\varcamparam)$ to its input parameters.
Consider an augmented version of the projection function $\varprojection^m(\varcamparam): \mathbb R^k \to \mathbb R^{2m}$, which is the concatenation of $\varprojection$ evaluated on $m$ points $\{\varpoint_j\}_{j=1}^m$. This set of points can be considered a \emph{proxy} for the 3D scene content being reconstructed; we want to measure how the 2D projections of these points change as a function of each camera parameter.

At any given point $\mat{\phi}^0$ in the camera parameter space, the effect of each camera parameter on each of the projected points $\varpixel_j = \varprojection(\varpoint_j; \varcamparam)$ can be represented by the Jacobian matrix
\begin{align}
    \left. \frac{d \varprojection^m}{d \varcamparam} \right|_{\varcamparam = \mat{\phi}^0} = \varjac_\varprojection \in \mathbb R^{2m \times k} \, .
\end{align}
The matrix $\mat{\Sigma}_\varprojection = {\varjac_\varprojection}^\top \varjac_\varprojection \in \mathbb R^{k \times k}$ has $rs$-th entry equal to 
\begin{align}
    \sum_{\ell=1}^m \left(\frac{d \varpixel_\ell}{d \varcamparam[r]}\right)^\top \left(\frac{d \varpixel_\ell}{d \varcamparam[s]}\right) \, .
    \label{eqn:correlation_coeffs}
\end{align}
On the diagonal this tells us the average motion magnitude induced by varying the $r$-th camera parameter $\varcamparam[r]$, and off the diagonal this tells how closely correlated motion is between camera parameters $\varcamparam[r]$ and $\varcamparam[s]$.

Our goal is to find a linear preconditioning matrix $\mat P$ such that substituting $\varprecon \tilde \varcamparam = \varcamparam$ into the projection function $\varprojection^m(\varprecon \tilde \varcamparam ) = \tilde \varprojection^m(\tilde \varcamparam)$ results in $\mat{\Sigma}_{\tilde \varprojection} = {\varjac_{\tilde\varprojection}}^\top \varjac_{\tilde\varprojection} $ that is equal to the identity matrix $\mat{I}_k$, where $\varjac_{\tilde\varprojection} = \varjac_\varprojection \varprecon$.
There are infinitely many solutions that satisfy the condition above, and we choose:
\begin{align}
    \varprecon = {\mat{\Sigma}_\varprojection}^{-\onehalf} = ({\varjac_\varprojection}^\top \varjac_\varprojection)^{-\onehalf} \,
\end{align} which is also known as the Zero Component Analysis (ZCA) whitening transform~\cite{kessy2018optimal}.

\figref{fig:overview} provides an overview of our complete method.
We visualize the process of preconditioning in \figref{fig:preconditioning}, including the covariance matrix $\varcov_\varprojection$ (b), the motion trails for the preconditioned camera (c) and the preconditioning matrix (d). Note how the preconditioner increases the magnitude of the focal length (first parameter), while  decreasing the magnitude of $y$-translation (sixth parameter), which exhibits the most motion in \figref{fig:se3focal_motion_trails}, and that the motion trails of the preconditioned cameras exhibit more similar motion magnitudes. 

Our preconditioner is only optimal with respect to the proxy-projection objective for the initial estimate of camera parameters $\varcamparam^0$. During optimization, the camera estimate deviates from $\varcamparam^0$, rendering it suboptimal. One could consider dynamically reparameterizing the preconditioned camera parameters with an updated optimal preconditioner, but this would render the running averages of the gradient moments used in optimizers like Adam~\cite{kingma2014adam} inaccurate. In practice, we observe that computing the preconditioner at initialization works sufficiently well.

Our analysis above assumes that cameras are parameterized independently from one another. However, in many reconstruction settings all images are taken with the same camera and lens, and thus their intrinsic parameters are shared during optimization. We can account for this case by imposing an additional loss term $\mathcal{L}_{\text{shared}}$ that minimizes the variance of the shared intrinsic parameters.

In practice, we add a small dampening parameter to the diagonals of the covariance matrix $\varcov_\varprojection$ in order to better condition the matrix before taking the inverse matrix square root:
\begin{align}
    \varprecon &= \left( \varcov_\Pi + \lambda \diag(\varcov_\Pi) + \mu \mat{I} \right)^{-\onehalf}\,,
\end{align}
where $\lambda$ and $\mu$ are hyperparameters. Note that this is especially important when including lens distortion parameters, where very small changes may dramatically affect the point projection.

\subsection{Implementation Details}

\paragraph{Model parameters}
We implement our method on top of Zip-NeRF~\cite{barron2023zip}. 
We adopt a coarse-to-fine strategy which progressively increases the contribution of higher resolution NGP grids.
Following \citet{heo2023robust}, we modulate the contribution of each grid by adjusting the learning rate of the grid parameters rather than by scaling the features, which side-steps this issue.
We use the Adam~\cite{kingma2014adam} optimizer for both the model and camera parameters. For the model, we use the same optimizer parameters as in Zip-NeRF, while for the camera parameters we use a learning rate that logarithmically decays from $10^{-3}$ to $10^{-4}$ over training. As with the scene model, the camera learning rate is warmed up using a multiplier that is cosine-decayed from $10^{-8}$ to 1 for the first \num[group-separator={,}]{2500}
 iterations. As in \citet{barron2023zip}, we train on 8 NVIDIA V100 GPUs for \num[group-separator={,}]{25000} steps, which takes a training time of approximately 2 hours.
For preconditioning we use the dampening parameters $\lambda=10^{-1}$ and $\mu=10^{-8}$. For scenes with shared intrinsics (360, perturbed 360), we weight the variance losses of the focal length by $10^{-1}$, the principal point by $10^{-2}$, and the radial distortion by $10^{-2}$. We do not use these losses for the NeRF-Synthetic dataset.

\paragraph{Camera parameterizations}
All camera parameterizations are implemented as residuals $\vardeltacamparam_i$ that are applied to the initial camera parameters $\varcamparam_i^0$. We use the following camera parameterizations in our experiments, more details are provided in the supplementary.
\begin{itemize}
    \item  \camerafont{SCNeRF}: We use the parameterization of SCNeRF~\cite{jeong2021self}, which uses a 6D rotation~\cite{zhou2019continuity}, a residual on the 3D translation ($t_x,t_y,t_z$), the focal lengths ($f_u,f_v$), the principal point ($u_0,v_0$), and the two radial distortion coefficients ($k_1,k_2$). We omit the raxel parameters and do not use the projected ray loss proposed in SCNeRF.
    
    \item \camerafont{SE3}: We use an $\sethree$ camera parameterization as in BARF~\cite{lin2021barf}. This corresponds to an improved version of BARF that has been adapted to the instant NGP setting.

    \item \camerafont{SE3+Focal+Intrinsics}: We augment the \camerafont{SE3} formulation with additional focal length, principal point, and two radial lens distortion parameters. The focal length residual is parameterized as a multiplicative log scale such that $f' = f\exp{\Delta f}$. In the evaluation section, we consider ablations labeled by \camerafontnobf{SE3+Focal} and \camerafontnobf{SE3+Focal+PrincipalPoint}, and \camerafontnobf{SE3} that does not optimize over intrinsic parameters.

    \item \camerafont{FocalPose+Intrinsics}: We use the joint pose and focal length parameterization proposed in FocalPose~\cite{ponimatkin2022focalpose}, which is designed for object-centric representations.  Similar to the SE3 parameterization, we add principal point and two radial lens distortion parameters. We note that FocalPose was proposed for an iterative pose estimation, and we thus modify it by reparameterizing it as a residual with respect to the initial estimates instead of the previous estimate.
\end{itemize}

\paragraph{Point sampling for preconditioning} 
To compute the preconditioner for a camera, we sample points in its view frustum, where pixel positions $\varpixel=(u,v)$ are sampled uniformly and normalized ray distances $d\in[0,1]$ are sampled uniformly between a normalized near and far plane $d\sim\mathcal{U}(n,f)$. The sampled depths are unnormalized using the power ladder transform of Zip-NeRF~\cite{barron2023zip}, with a curve identical to its ray sampling distribution.

\section{Results}

We evaluate our model on several real and synthetic datasets. Please see our supplemental video for additional comparisons and results.\\

\paragraph{
360
\protect\footnote{\label{360dataset}We use the original images that have not been undistorted by COLMAP, whereas mip-NeRF 360 and Zip-NeRF \emph{do} use the undistorted images, resulting in slight differences between the metrics reported in our tables and theirs.}
:
}
We use the dataset of 4 indoor and 5 outdoor scenes from mip-NeRF 360~\cite{barron2022mipnerf360}, with 100 to 300 views each. We use the same evaluation protocol as mip-NeRF 360. Instead of pre-undistorting images, we directly use the original images while accounting for the camera distortion on the fly when casting rays, avoiding a lossy image resampling step. Our model is trained and evaluated on images with resolution $1560\times1040$ for indoor scenes and $1228\times816$ for outdoor scenes.

\paragraph{Perturbed-360:}
In order to gauge the robustness of our method to camera errors, we evaluate on a version of the 360 dataset with perturbed cameras. We perturb the position and orientation of the camera via the look-at point and position of the camera with $\mathcal{N}(0, 0.005)$ and $\mathcal{N}(0, 0.005)$ respectively. We perturb the focal length with a random scale $\exp\mathcal{N}(0, \log(1.02))$ and perturb the distance to the origin (scene center) and the focal length with an identical random scale factor $\exp\mathcal{N}(0, \log(1.05))$ (akin to a dolly zoom). Lastly, we set the initial distortion parameters of all cameras to zero.
\paragraph{Perturbed-Synthetic:}
To compare the estimated camera parameters with ground truth camera parameters, we use the NeRF-Synthetic dataset proposed by \cite{mildenhall2020nerf} containing 8 objects. Each object is rendered from 100 viewpoints for training and another 200 different viewpoints for computing test error metrics.
We perturb the training cameras using the same strategy as the 360 perturbed dataset. We perturb both the camera look-at point and position with $\mathcal{N}(0, 0.1)$.
We perturb the focal length with a random scale factor $\exp\mathcal{N}(0, \log(1.2))$ and perturb the distance to the origin (scene center) and the focal length with an identical random scale factor $\exp\mathcal{N}(0, \log(1.1))$. We disable distortion for the synthetic dataset. Note that we use much stronger perturbations for the synthetic dataset since objects exhibit weaker perspective distortion compared to unbounded scenes. The strong perturbations make this dataset incredibly challenging.

\subsection{Evaluation}

When refining camera parameters, we recover the scene structure up to a similarity transform. In order to perform a quantitative evaluation with the test cameras, we follow the protocol of BARF and estimate the camera parameters of the test cameras using photometric test-time optimization. We omit the Procrustes alignment method proposed in BARF, as it only takes into account the camera positions, and breaks when the focal length of a camera changes significantly.
For all methods, we use the \camerafont{FocalPose+Intrinsics} parameterization with preconditioning and optimize using Adam~\cite{kingma2014adam} with a learning rate of 0.001 for 100 steps.

\begin{table}[t]
\centering

\caption{
Performance on the 360 dataset\protect\footnoteref{360dataset}.
}
\label{tab:non_perturbed_360}
\vspace{-0.1in}

\resizebox{0.8\linewidth}{!}{

\begin{tabular}{lc|ccc}
Camera Param. & \namehandle & PSNR \textuparrow & SSIM \textuparrow & LPIPS \textdownarrow \\
\hline

No camera opt. &                  &                   28.27 &                   0.825 &                   0.196 \\
\hline
SE3 &                             &                   28.48 &                   0.830 &                   0.191 \\
SE3 & \checkmark                  &                   28.49 &                   0.831 &                   0.191 \\
\hline
SCNeRF &                          &                   27.60 &                   0.782 &                   0.198 \\
SCNeRF & \checkmark               &                   28.52 &                   0.828 & \cellcolor{myyellow}0.184 \\
\hline
SE3+Focal+Intrinsics &            &                   28.73 & \cellcolor{myyellow}0.841 & \cellcolor{myorange}0.183 \\
SE3+Focal+Intrinsics & \checkmark & \cellcolor{myorange}28.79 & \cellcolor{myorange}0.842 &    \cellcolor{myred}0.182 \\
\hline
FocalPose+Intrinsics &            & \cellcolor{myyellow}28.76 & \cellcolor{myorange}0.842 &    \cellcolor{myred}0.182 \\
FocalPose+Intrinsics & \checkmark &    \cellcolor{myred}28.86 &    \cellcolor{myred}0.843 &    \cellcolor{myred}0.182
\end{tabular}
}
\end{table}

\tabref{tab:non_perturbed_360} shows that camera-optimizing NeRFs improve over our non-optimizing baseline, and that our preconditioned methods perform slightly better than the non-preconditioned alternatives except for the \camerafont{SE3} parameterization. We show a qualitative comparison in~\figref{fig:teaser} showcasing the differences in a challenging outdoor scene from the 360 dataset. All methods in this case are initialized with camera parameters estimated by using the COLMAP structure-from-motion software~\cite{schoenberger2016sfm}, showing that our camera-optimizing NeRF methods can improve upon COLMAP's camera estimates to achieve lower photometric error.

The differences between the variants become starker in the perturbed 360 dataset in ~\tabref{tab:perturbed_360_camera_image}, which is significantly harder than the unperturbed setting. In this case, camera optimization improves the results and camera metrics significantly, and the effects of preconditioning are more significant, improving the PSNR between $4.3$ and $6.8$dB for all camera parameterizations. We visualize the effects of camera preconditioning in \figref{fig:qualitative_on_perturbed}.

We observe similarly big improvements on the perturbed NeRF-Synthetic dataset, for which camera parameters were perturbed by much larger amounts than the 360 dataset. \tabref{tab:perturbed_synthetic_nerf_camera} and \figref{fig:qualitative_blender_perturbed} show that preconditioning improves all camera parameterizations across all metrics. It also shows that the \camerafont{FocalPose} parameterization significantly outperforms other alternatives considered. This trend extends to the 360 scenes. This is noteworthy, as FocalPose was proposed in the context of 3D object pose estimation and we are not aware of previous works that use it for optimizing camera parameters in an inverse rendering approach.

In \tabref{tab:perturbed_360_ablation_diag}, we compare to a variant of our method that uses per-parameter scaling factors, i.e., a diagonal version of our preconditioner shown in Eqn. \ref{eqn:correlation_coeffs}. This comparison shows that modeling the correlations is helpful for optimization and that a full matrix preconditioner consistently outperforms a diagonal one.

\begin{table}[t]
\centering

\caption{
Performance on the perturbed 360 dataset, using COLMAP estimated parameters as ground truth. $R$ is camera orientation error (in degrees), $\mat{p}$ is position error (in world units), $f$ is focal length error (in pixels).
\label{tab:perturbed_360_camera_image}
\vspace{-0.1in}
}
\centering
\resizebox{\columnwidth}{!}{
\begin{tabular}{l@{}c|ccc|ccc}
& \multicolumn{1}{c|}{\,} & \multicolumn{3}{c|}{Image Metrics} & \multicolumn{3}{c}{Camera Metrics} \\
Camera Param. & \namehandle & PSNR\textuparrow & SSIM\textuparrow & LPIPS\textdownarrow & $R$\textdownarrow & $\mat{p}$\textdownarrow & $f$\textdownarrow\\
\hline

No camera opt. &                  &                   18.50 &                   0.403 &                   0.545 &  0.565  & 0.045 & 55.7  \\
\hline
SCNeRF &                          &                   25.13 &                   0.658 &                   0.319 &                   0.672 &                   0.047 &                   76.1 \\
SCNeRF & \checkmark               &                   26.57 &                   0.742 &                   0.226 &                   0.660 &                   0.028 & \cellcolor{myyellow}24.0 \\
\hline
SE3+Focal+Intrinsics &            &                   27.59 &                   0.796 & \cellcolor{myyellow}0.218 & \cellcolor{myyellow}0.448 & \cellcolor{myyellow}0.024 &                   36.9 \\
SE3+Focal+Intrinsics & \checkmark &    \cellcolor{myred}28.22 & \cellcolor{myorange}0.822 & \cellcolor{myorange}0.202 & \cellcolor{myorange}0.432 & \cellcolor{myorange}0.022 &    \cellcolor{myred}21.1 \\
\hline
FocalPose+Intrinsics &            & \cellcolor{myyellow}27.60 & \cellcolor{myyellow}0.797 &                   0.219 &                   0.508 &                   0.026 &                   38.8 \\
FocalPose+Intrinsics & \checkmark & \cellcolor{myorange}28.14 &    \cellcolor{myred}0.823 &    \cellcolor{myred}0.201 &    \cellcolor{myred}0.426 &    \cellcolor{myred}0.016 & \cellcolor{myorange}21.9

\end{tabular}
}

\end{table}

\begin{table}[t]
\centering

\caption{
Performance on the perturbed synthetic dataset, where ground truth parameters are known. $R$ is camera orientation error (in degrees), $\mat{p}$ is position error (in world units), $f$ is focal length error (in pixels).
\label{tab:perturbed_synthetic_nerf_camera}
\vspace{-0.1in}
}
\centering
\resizebox{\columnwidth}{!}{
\begin{tabular}{lc|ccc|ccc}
& \multicolumn{1}{c|}{\,} & \multicolumn{3}{c|}{Image Metrics} & \multicolumn{3}{c}{Camera Metrics} \\
Camera Param. & \namehandle & PSNR\textuparrow & SSIM\textuparrow & LPIPS\textdownarrow & $R$\textdownarrow & $\mat{p}$\textdownarrow & $f$\textdownarrow\\
\hline

No camera opt. &       &                   14.92 &                   0.734 &                   0.284 &  1.640 & 0.691 & 204.4  \\
\hline
SE3 &                  &                   18.96 &                   0.813 &                   0.185 &                   1.382 &                   0.675 &                   204.4 \\
SE3 & \checkmark       &                   21.37 &                   0.857 &                   0.136 &                   0.957 &                   0.674 &                   204.4 \\
\hline
SE3+Focal &            &                   21.97 &                   0.866 &                   0.126 &                   0.828 &                   0.588 &                   177.5 \\
SE3+Focal & \checkmark & \cellcolor{myyellow}25.98 & \cellcolor{myyellow}0.913 & \cellcolor{myyellow}0.079 & \cellcolor{myorange}0.487 & \cellcolor{myyellow}0.287 & \cellcolor{myyellow}87.3 \\
\hline
FocalPose &            & \cellcolor{myorange}26.80 & \cellcolor{myorange}0.918 & \cellcolor{myorange}0.074 & \cellcolor{myyellow}0.572 & \cellcolor{myorange}0.245 & \cellcolor{myorange}68.9 \\
FocalPose & \checkmark &    \cellcolor{myred}28.29 &    \cellcolor{myred}0.930 &    \cellcolor{myred}0.064 &    \cellcolor{myred}0.420 &    \cellcolor{myred}0.186 &    \cellcolor{myred}54.3

\end{tabular}
}

\end{table}

\begin{table}[t]
\centering
\caption{
Effects of using a diagonal preconditioner vs. our full matrix preconditioner on the perturbed 360 dataset.
\label{tab:perturbed_360_ablation_diag}
}
\vspace{-1em}
\resizebox{\columnwidth}{!}{
\begin{tabular}{@{}lc|ccc|ccc@{}}
& \multicolumn{1}{c|}{\,} & \multicolumn{3}{c|}{Image Metrics} & \multicolumn{3}{c}{Camera Metrics} \\
Camera Param. & \namehandle & PSNR\textuparrow & SSIM\textuparrow & LPIPS\textdownarrow & $R$ (\textdegree)\textdownarrow & $\mat{p}$\textdownarrow & $f$ (px)\textdownarrow\\
\hline
SCNeRF & diag                &                   25.96 &                   0.718 &                   0.243 &                   0.707 &                   0.025 &                   34.0 \\
SCNeRF & full                &                   26.61 &                   0.745 & \cellcolor{myyellow}0.225 &                   0.660 &                   0.028 & \cellcolor{myyellow}23.7 \\
\hline
SE3+Focal+Intrinsics & diag  &                   27.55 &                   0.799 & \cellcolor{myorange}0.218 & \cellcolor{myyellow}0.465 & \cellcolor{myyellow}0.021 &                   38.9 \\
SE3+Focal+Intrinsics & full  &    \cellcolor{myred}28.18 & \cellcolor{myorange}0.821 &    \cellcolor{myred}0.202 &    \cellcolor{myred}0.432 &                   0.022 &    \cellcolor{myred}21.1 \\
\hline
FocalPose+Intrinsics & diag  & \cellcolor{myyellow}27.56 & \cellcolor{myyellow}0.800 & \cellcolor{myorange}0.218 &                   0.508 & \cellcolor{myorange}0.020 &                   40.5 \\
FocalPose+Intrinsics & full  & \cellcolor{myorange}28.13 &    \cellcolor{myred}0.822 &    \cellcolor{myred}0.202 & \cellcolor{myorange}0.433 &    \cellcolor{myred}0.016 & \cellcolor{myorange}21.9
\end{tabular}
}
\end{table}

\begin{table}[t]
\centering

\caption{
Performance on the ARKit dataset.
\label{tab:arkit}
}
\vspace{-0.1in}

\begin{tabular}{l@{}c|ccc}
& \namehandle & PSNR\textuparrow & SSIM\textuparrow & LPIPS\textdownarrow \\
\hline

No camera opt. &                  &                   21.12 &                   0.615 &                   0.432 \\
\hline
FocalPose &                       &                   25.50 &                   0.796 & \cellcolor{myyellow}0.270 \\
FocalPose & \checkmark            & \cellcolor{myyellow}25.53 & \cellcolor{myyellow}0.797 & \cellcolor{myyellow}0.270 \\
\hline
FocalPose+Intrinsics &            & \cellcolor{myorange}25.74 & \cellcolor{myorange}0.808 & \cellcolor{myorange}0.260 \\
FocalPose+Intrinsics & \checkmark &    \cellcolor{myred}25.97 &    \cellcolor{myred}0.817 &    \cellcolor{myred}0.258

\end{tabular}
\end{table}

\fboxsep=0pt %
\fboxrule=0.4pt %

\newcommand{\textimageperturbed}[4]{
	\begin{overpic}[width=\columnwidth]{#1}
	\put (-1.0,0.0) {\sethlcolor{white}\footnotesize\hl{$#2$}}
	\put (-1.0, 83.0) {\sethlcolor{white}\footnotesize\hl{$#3$}}
	\put (-1.0, 165.0) {\sethlcolor{white}\footnotesize\hl{$#4$}}
    \end{overpic}
    \vspace{-1.2em}
    }

\def\figcelltt#1#2#3#4#5#6{\begin{subfigure}{#1\columnwidth}\centering\textimageperturbed{#2}{#3}{#4}{#5}
\def\temp{#6}\ifx\temp\empty\else\caption*{#6}\fi\end{subfigure}}

\begin{figure}
	\captionsetup[sub]{labelformat=parens}
    \begingroup
       \captionsetup{type=figure}
        \figcelltt{0.49}{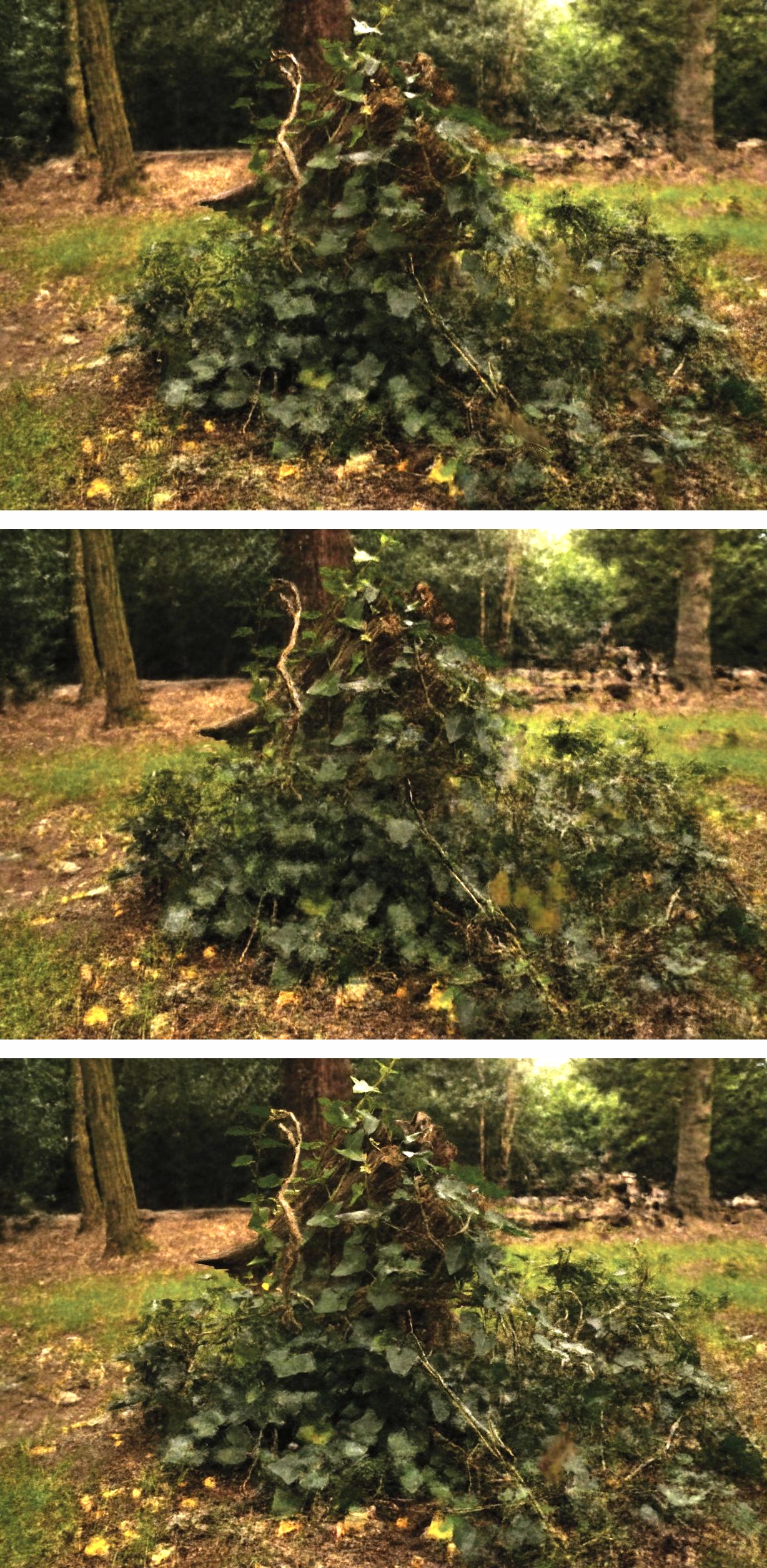}{\mathrm{FocalPose+Intrinsics}}{\mathrm{SE3+Intrinsics}}{\mathrm{SCNeRF}}{Without \nameacronym}
        \figcelltt{0.49}{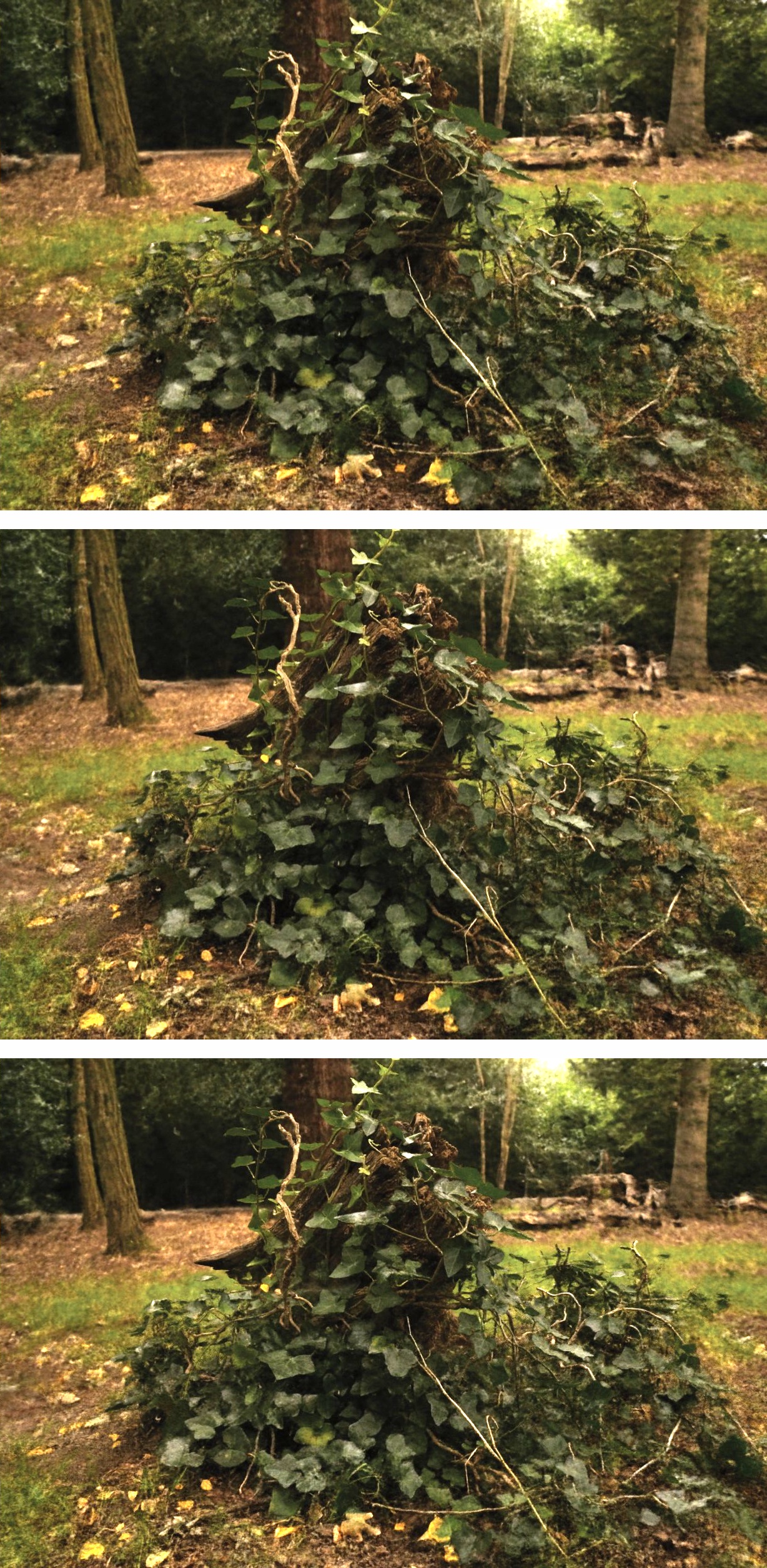}{}{}{}{With \namehandle}
    \endgroup
    \caption{
    Comparison of different camera parameterizations with and without preconditioning on the perturbed 360 dataset stump scene. Without preconditioning (left) the scene appears to be blurry whereas preconditioning (right) yields to significantly sharper results.
    }
    \label{fig:qualitative_on_perturbed}

\end{figure}

\subsection{Cellphone Captures}
\begin{figure*}
	\captionsetup[sub]{labelformat=parens}
	
    \begin{subfigure}[b]{1\textwidth}
        \centering
           	\figcellt{0.32}{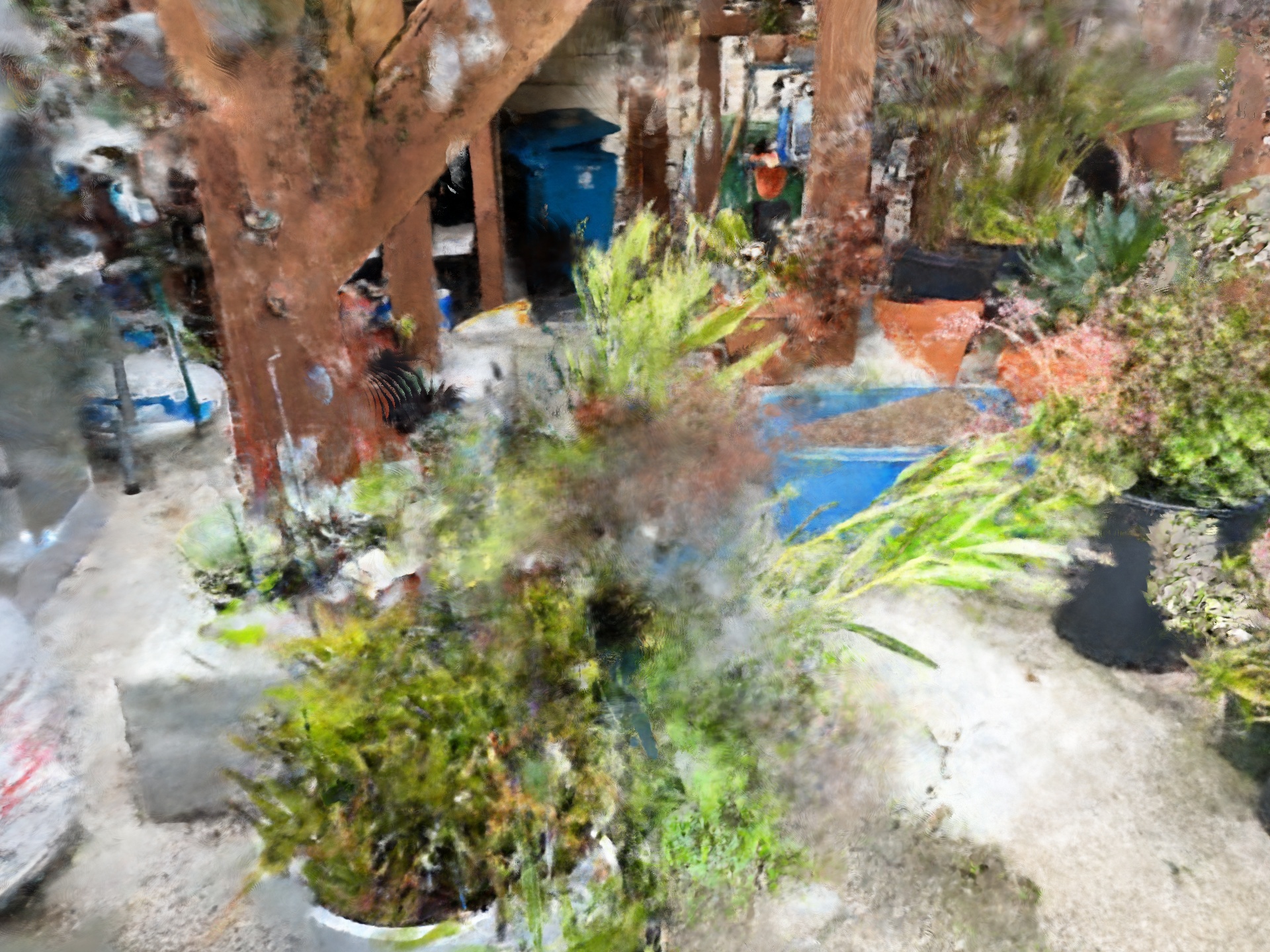}{}{clip,trim=400 50 500 50}\hspace{-1.0pt}
           	\figcellt{0.32}{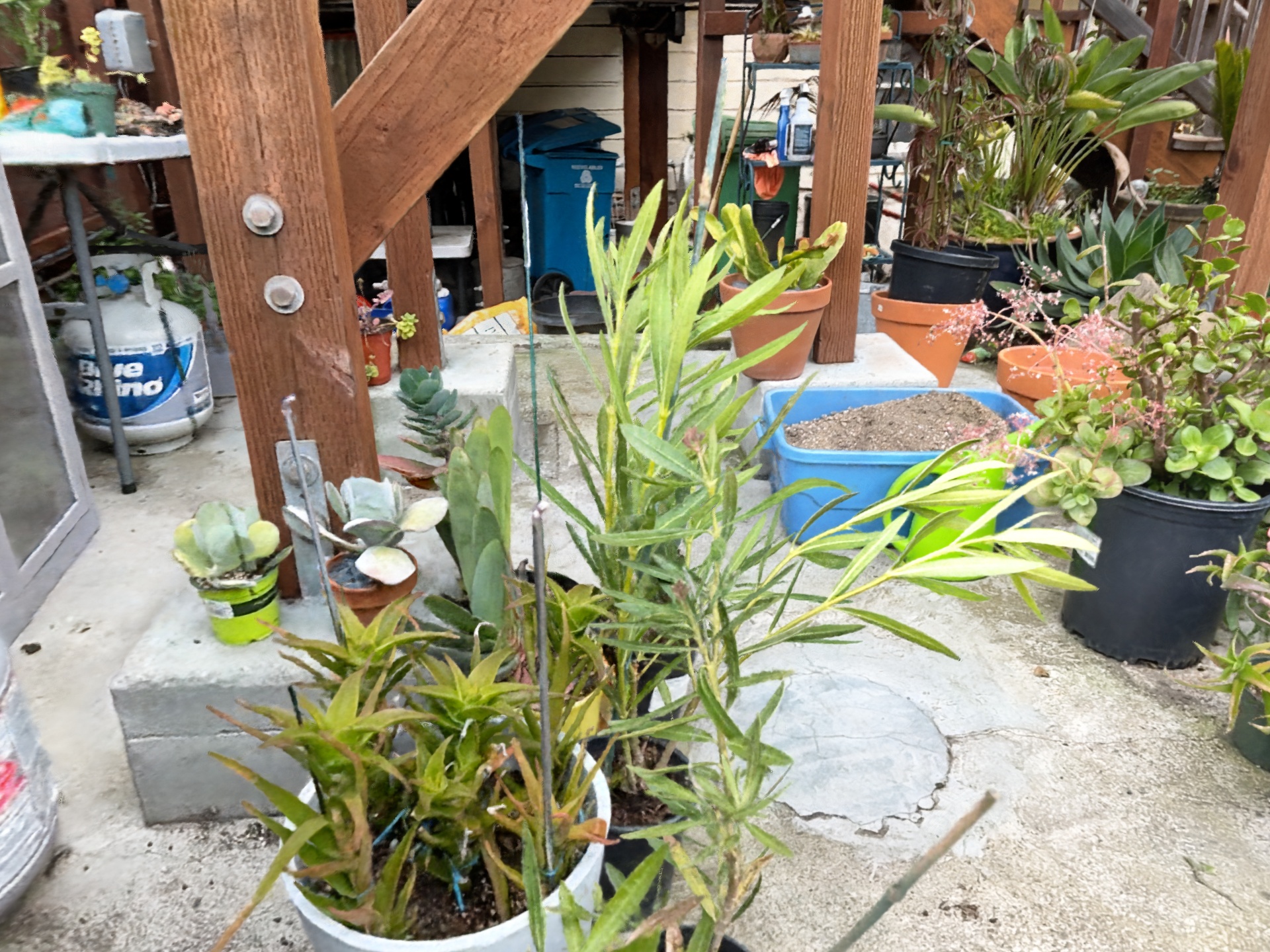}{}{clip,trim=400 50 500 50}\hspace{-1.0pt}
           	\figcellt{0.32}{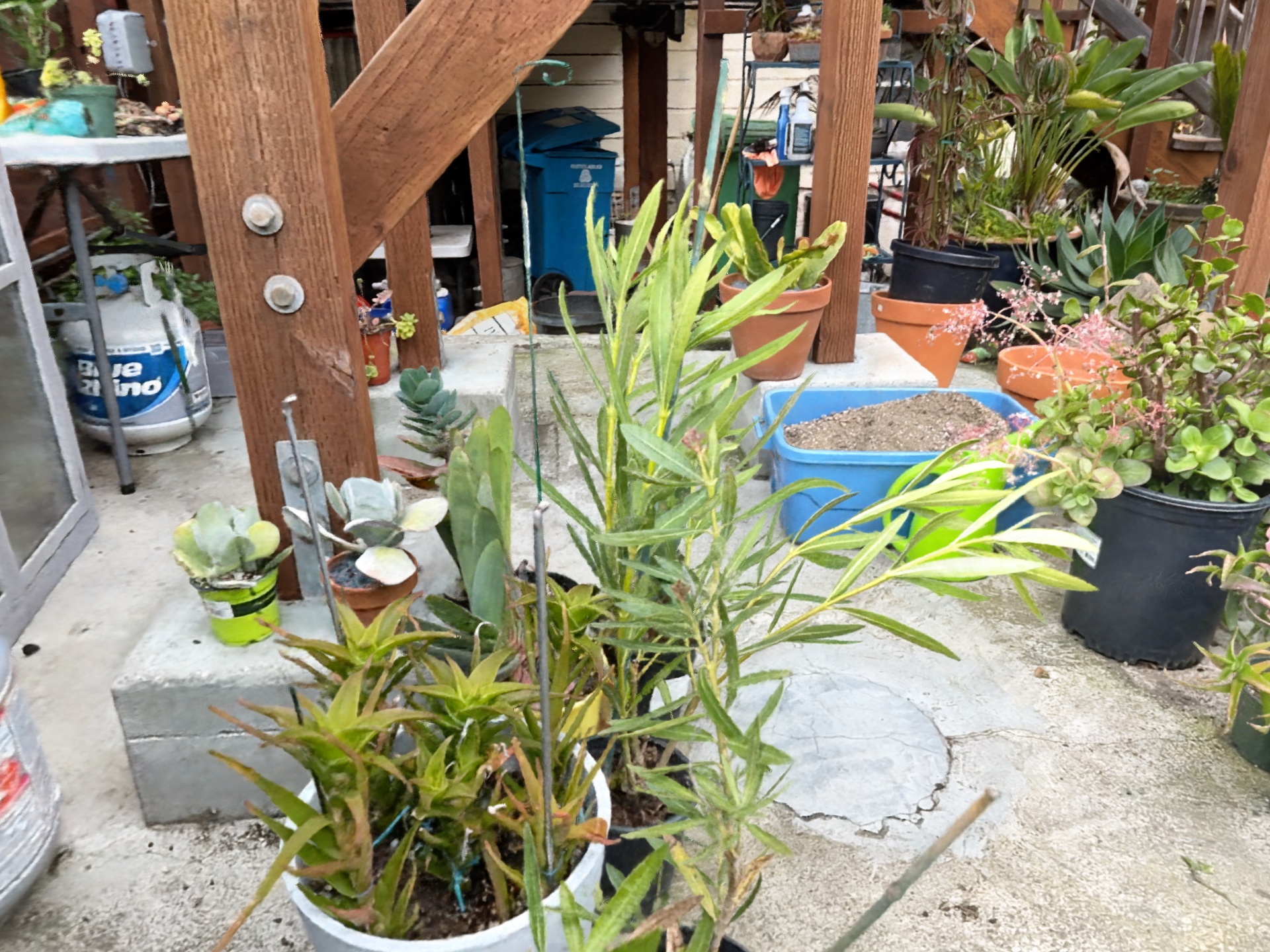}{}{clip,trim=400 50 500 50}\hspace{-1.0pt}
    \end{subfigure}
    
    \begin{subfigure}[b]{1\textwidth}
        \centering
           	\figcellt{0.32}{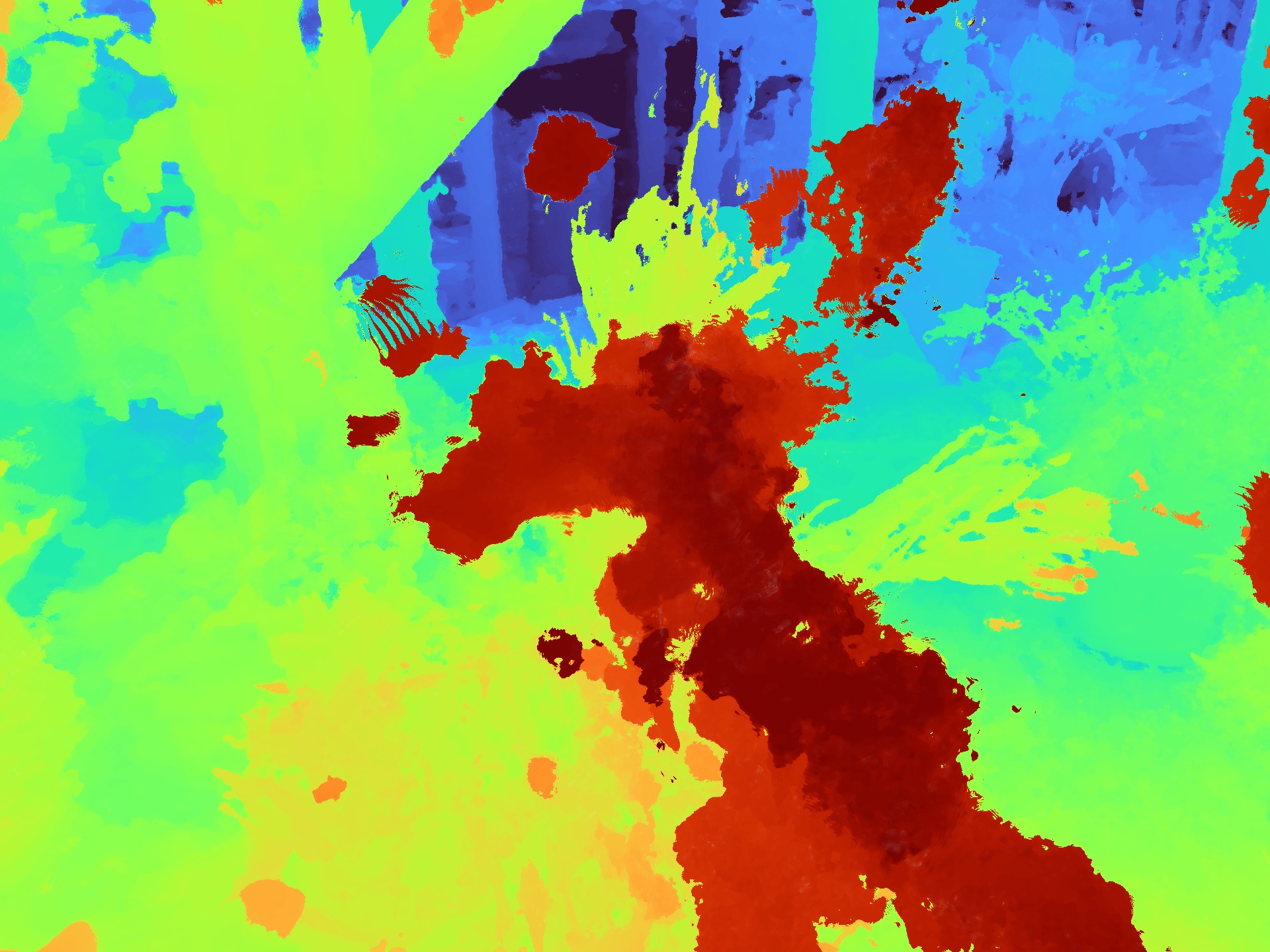}{No Camera Optimization}{clip,trim=400 50 500 50}\hspace{-1.0pt}
           	\figcellt{0.32}{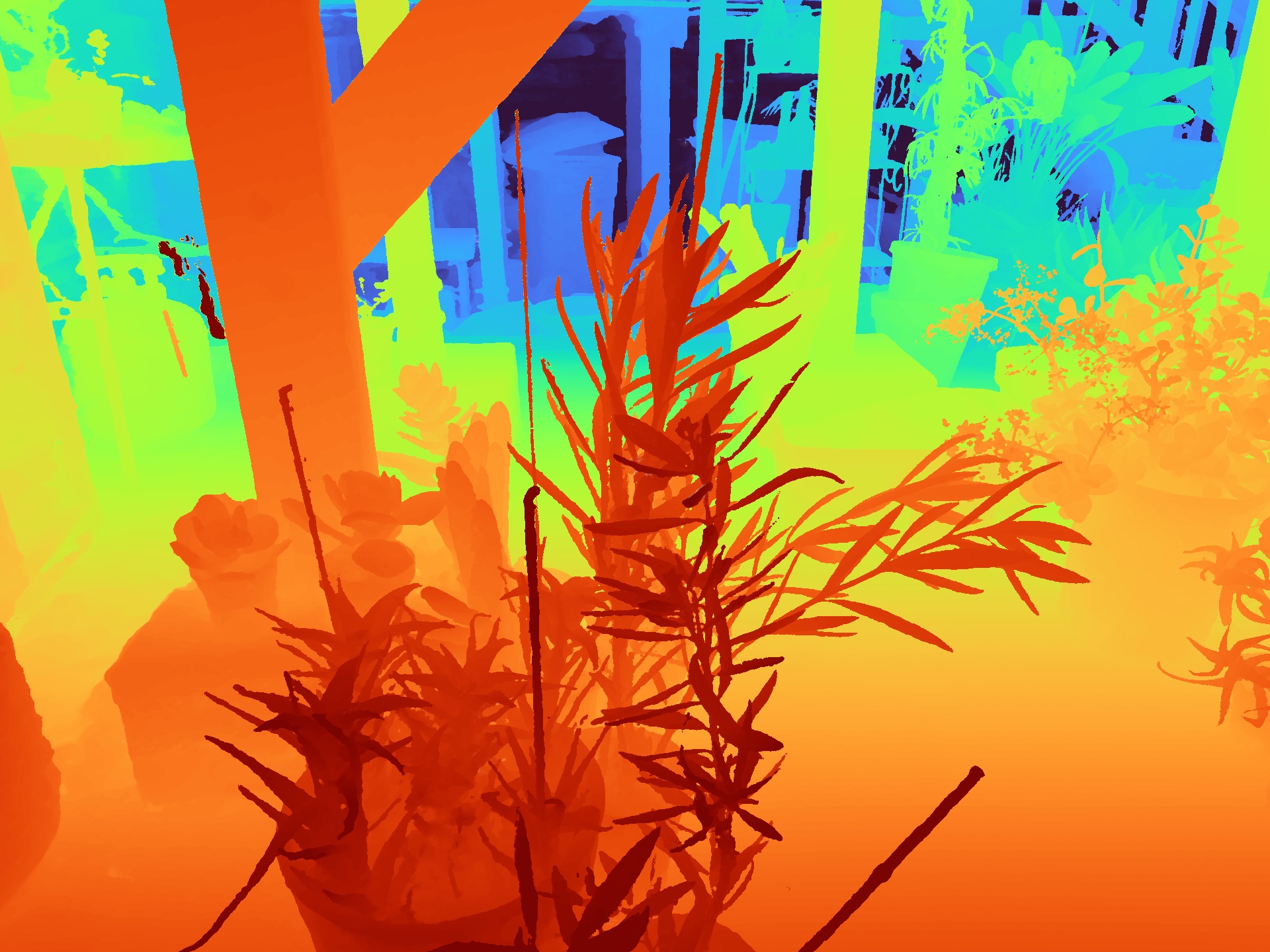}{+ Camera Optimization}{clip,trim=400 50 500 50}\hspace{-1.0pt}
           	\figcellt{0.32}{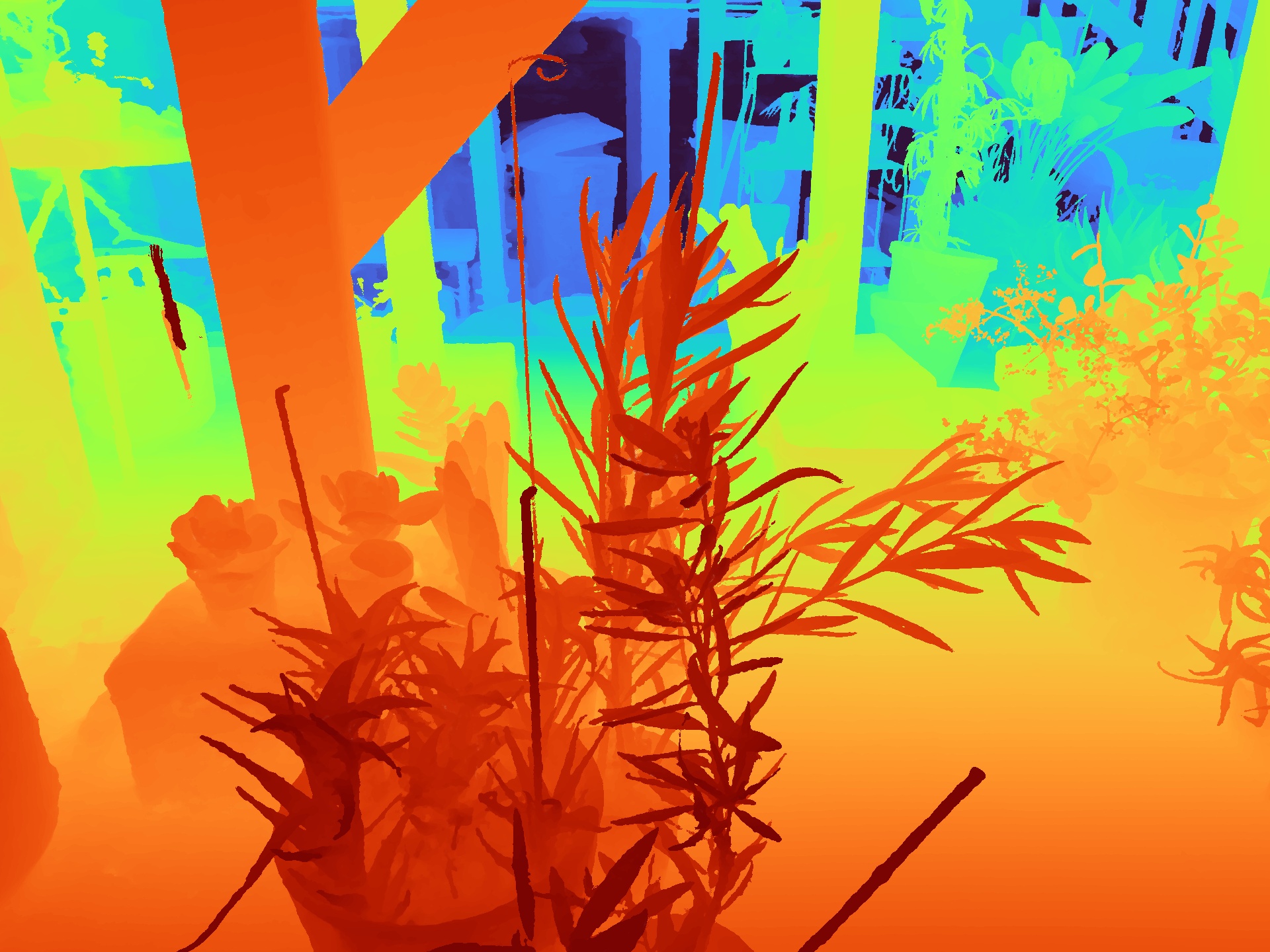}{+ \namehandle}{clip,trim=400 50 500 50}\hspace{-1.0pt}
    \end{subfigure}
    
    \caption{
    Rendered images and depth maps of a test viewpoint for a scene captured with ARKit. The variant without pose optimization is not able to recover the structure of the scene. Additionally, the pole in the top-middle of the frame is only recovered only by the preconditioning method.    }
    \label{fig:arkit}
\end{figure*}

Modern cellphones are able to estimate pose information using visual-inertial odometry, which is often used for Augmented Reality (AR) applications. While the poses are sufficiently good for AR effects, it can be challenging to recover high quality NeRFs from those sequences without resorting to running expensive, offline SfM pipelines. We evaluate the performance of our method using the promising \camerafont{FocalPose+Intrinsics} camera parameterization on a dataset of 9 casually captured scenes using the open source NeRF Capture app\footnote{\url{https://github.com/jc211/NeRFCapture}} on an iPhone 13 Pro, which exports camera poses estimated by ARKit. We report quantitative metrics in \tabref{tab:arkit} and show qualitative comparisons in \figref{fig:arkit} that highlight the benefits of camera optimization.

\section{Limitations and Future Work}

Similar to previous work, our preconditioning approach does not always prevent the optimization from falling into local minima.
For example, the white floaters in the synthetic NeRF dataset are caused by the model blocking parts of the scene that are not aligned due to bad camera poses. 
Our preconditioner can be improved if the distribution of points in the scene are no a priori. For example, feature points from SfM methods may be used to compute the preconditioner. Our method may also benefit from dynamically updating the preconditioner as the cameras are optimized, both by recomputing the preconditioner at the current parameter estimate and by resampling the points based on the current trained geometry of the scene.
Finally, understanding how the preconditioners interact with adaptive optimizers like Adam will likely lead to insights that allow for designing better optimizers for camera-optimizing NeRFs.

\section{Conclusion}
We proposed a simple yet effective way of preconditioning camera parameters for NeRF optimization. We study existing camera parameterizations and observe that their parameters can be (a) correlated and (b) have different units, e.g. translation in world units vs focal length in pixels. Both lead to undesirable optimization behavior. To address these issues, we propose a novel preconditioner for the pose optimization problem that decorrelates the camera parameters and normalizes their effects on the rendered images. We demonstrate the efficacy of our approach in multiple settings, improving camera-optimizing NeRFs for various camera parameterizations.

\begin{anonsuppress}
\section*{Acknowledgments}
We thank Rick Szeliski and Sameer Agarwal for their comments on the text; and Ben Poole, Aleksander Holynski, Pratul Srinivasan, Ben Attal, Peter Hedman, Matthew Burrus, Laurie Zhang, Matthew Levine, and Forrester Cole for their advice and help.
\end{anonsuppress}

\def\figcelltwide#1#2#3#4{\begin{subfigure}{#1\textwidth}\centering\includegraphics[width=\textwidth,#4]{#2} \def\temp{#3}\ifx\temp\empty\else\caption*{#3}\fi\end{subfigure}}

\begin{figure*}
	\captionsetup[sub]{labelformat=parens}
    \captionsetup{type=figure}
   	\figcelltwide{0.17}{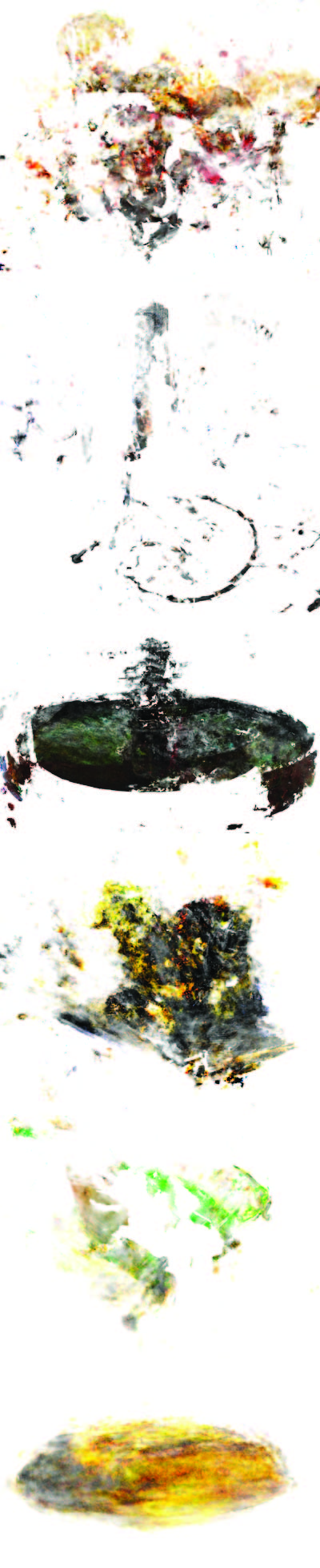}{\small No camera optimization}{}
   	\figcelltwide{0.17}{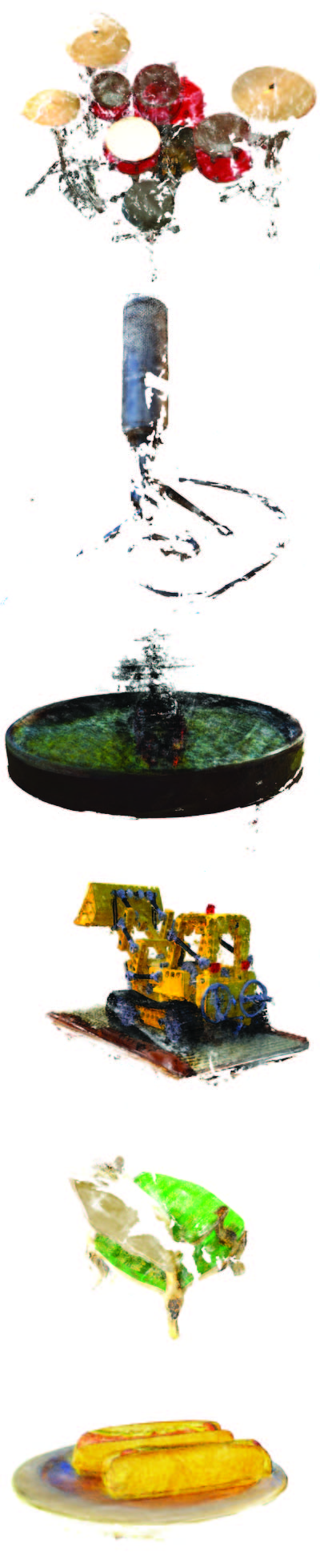}{\small \textbf{SE3}}{}
   	\figcelltwide{0.17}{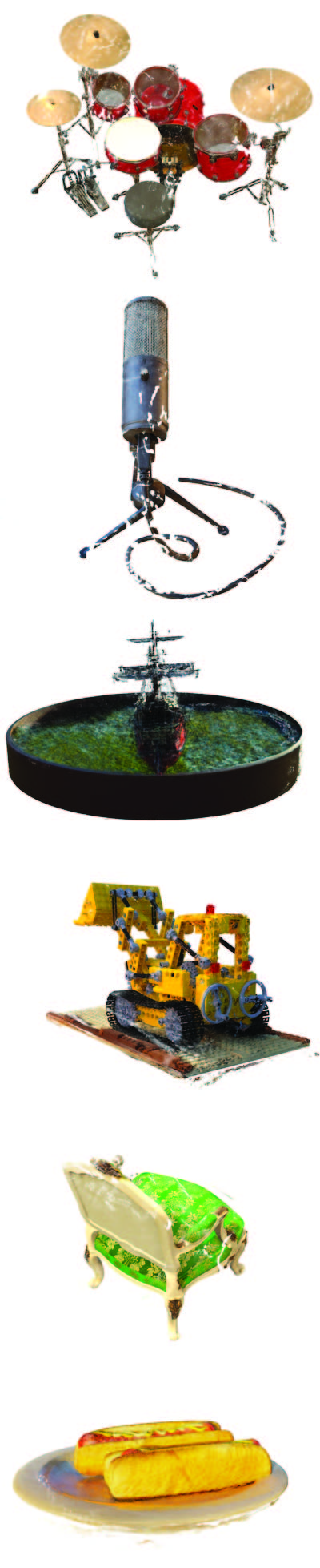}{\small \textbf{SE3+Focal}}{}
   	\figcelltwide{0.17}{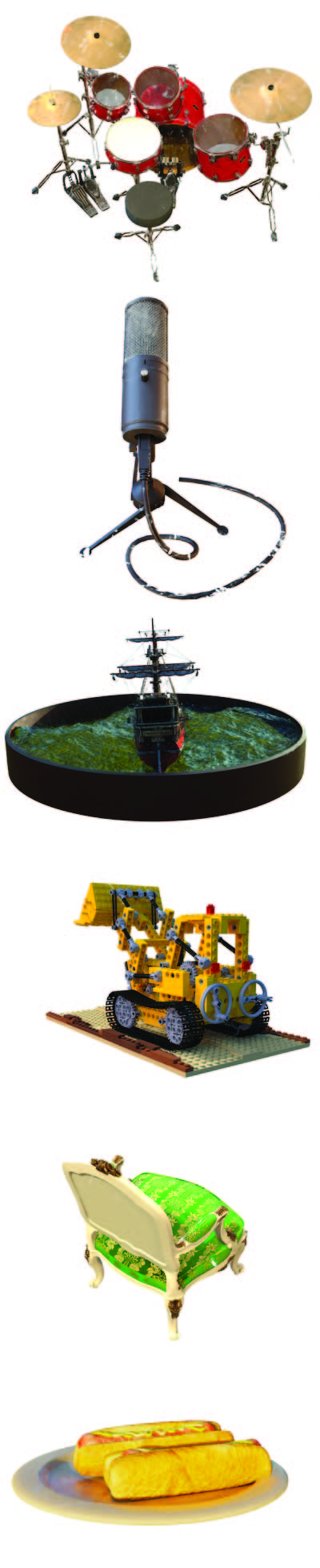}{\small  \textbf{SE3+Focal} with \namehandle}{}
   	\figcelltwide{0.17}{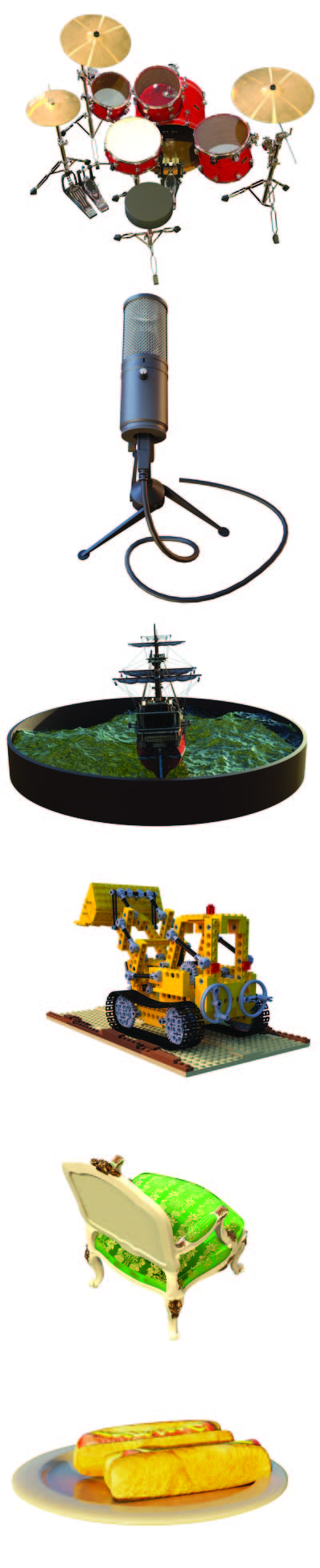}{\small ground truth}{}
    \caption{
    Qualitative comparisons on the perturbed synthetic dataset.
    }
    \vspace{0.5em}
    \label{fig:qualitative_blender_perturbed}
\end{figure*}
\fboxsep=0pt %
\fboxrule=0.4pt %

\begin{anonsuppress}
\appendix

\section{Camera Parameterizations}

Here we describe the camera parameterizations in detail. For each parameter we denote the initial parameters as the bare symbol $\phi$, the residual parameter as $\Delta\phi$, and the current estimate as $\phi'$.

\paragraph{\camerafont{SE3}} We use an $\sethree$ screw-axis parameterization $\mathcal{S} = (\logrot; \mat{v})\in\mathbb{R}^6$ to encode the rigid pose of the camera. This is identical to the one used by BARF~\cite{lin2021barf}.
\begin{align}
    \mathcal{S}' = \mathcal{S}^0 + \Delta\mathcal{S}
\end{align}

\paragraph{\camerafont{+Focal}} We can augment parameterizations with a focal length parameter. We use a multiplicative log residual.
\begin{align}
    f' = f \exp{\Delta f}
\end{align}
\paragraph{\camerafont{+Intrinsics}} Likewise, we augment camera parameterizations with a principal point ($u,v$) and 2nd order radial distortion~\cite{conrady1919decentred} residuals ($k_1,k_2$).
\begin{align}
    u_0' = u_0 + \Delta u_0 \\
    v_0' = v_0 + \Delta v_0 \\
    k_1' = k_1 + \Delta k_1 \\
    k_2' = k_2 + \Delta k_2
\end{align}

\paragraph{\camerafont{FocalPose}} We use the camera formulation of \citet{ponimatkin2022focalpose}, but adapt it to use with optimized residual parameters rather than sequential updates from a feed-forward network:
\begin{align}
    f' &= f \exp{\Delta f} \\
    z' &= z \exp{\Delta z} \\
    x' &= \left( \frac{\Delta x}{f} + \frac{x}{z} \right) z' \\
    y' &= \left( \frac{\Delta y}{f} + \frac{y}{z} \right) z' \,,
\end{align}
where $f$ is the focal length, and $x,y,z$ encode the translation of the camera with respect to the origin.

\paragraph{\camerafont{SCNeRF}} We use the SCNeRF camera parameterization, with the exception of the local raxel parameters that capture generic non-linear aberrations. SCNeRF uses the 6D continuous rotation parameterization $\mat{a} \in \mathbb{R}^6$ of \citet{zhou2019continuity}. Unlike the other models, \camerafont{SCNeRF} uses an additive bias as the residuals for the vertical and horizontal focal lengths as well as translation.
\begin{align}
    f_u' &= f_u + \Delta f_u \\
    f_v' &= f_v + \Delta f_v \\
    x' &= x + \Delta x \\
    y' &= y + \Delta y \\
    z' &= z + \Delta z \\
    \mat{a} &= \mat{a} + \Delta \mat{a}
\end{align}
In addition, SCNeRF includes the same principal point and radial distortion parameterization as \camerafont{Intrinsics}.

\subsection{Variants}
\begin{table}[t]
\centering

\caption{
Ablation on the outdoor scene of the perturbed 360 dataset showing the performance of different parameterization choices. Pix. Scale scales pixel units to world units by multiplying the focal length to the residual (e.g., the focal length, principal point, and $x,y$ translation parameters of FocalPose). Log Scale parameterizes length residuals as a log scale instead of a bias.
\label{tab:ablation_manual_reparam}
}
\vspace{-0.1in}

\resizebox{1.0\linewidth}{!}{
\setlength{\tabcolsep}{2pt}

\begin{tabular}{lccc|ccc|ccc}
&Pix. Scale & Log Scale & \namehandle & PSNR \textuparrow & SSIM \textuparrow & LPIPS \textdownarrow & $R$\textdownarrow & $\mat{p}$\textdownarrow & $f$\textdownarrow \\
\hline

FocalPose+Intrinsics & & &                                  &                   24.01 &                   0.614 &                   0.368 &                   0.756 &                   0.048 &                   76.0 \\
SE3+Focal+Intrinsics & & &                                  &                   25.53 &                   0.682 &                   0.312 &                   0.672 &                   0.049 &                   75.8 \\
FocalPose+Intrinsics & & \checkmark &                       &                   26.04 &                   0.714 &                   0.287 &                   0.528 &                   0.043 &                   55.8 \\
SE3+Focal+Intrinsics & & \checkmark &                       &                   27.56 &                   0.796 &                   0.219 &    \cellcolor{myred}0.368 &                   0.025 &                   37.7 \\
FocalPose+Intrinsics & \checkmark & &                       &                   27.57 &                   0.795 &                   0.219 &                   0.507 &                   0.024 &                   38.3 \\
SE3+Focal+Intrinsics & \checkmark & &                       &                   27.61 &                   0.797 &                   0.218 &                   0.451 &                   0.024 &                   37.0 \\
FocalPose+Intrinsics & \checkmark & \checkmark &            &                   27.60 &                   0.797 &                   0.218 &                   0.507 &                   0.026 &                   38.8 \\
SE3+Focal+Intrinsics & \checkmark & \checkmark &            &                   27.65 &                   0.798 &                   0.217 &                   0.446 &                   0.024 &                   37.0 \\
FocalPose+Intrinsics & & & \checkmark                       &                   27.66 &                   0.802 &                   0.214 &                   0.533 & \cellcolor{myyellow}0.021 &                   33.3 \\
FocalPose+Intrinsics & & \checkmark & \checkmark            &                   27.81 &                   0.808 &                   0.212 &                   0.515 & \cellcolor{myorange}0.020 &                   31.6 \\
SE3+Focal+Intrinsics & & \checkmark & \checkmark            &                   28.07 &                   0.819 &                   0.204 &                   0.458 &                   0.024 & \cellcolor{myorange}20.6 \\
SE3+Focal+Intrinsics & & & \checkmark                       &                   28.08 &                   0.820 & \cellcolor{myyellow}0.203 &                   0.462 &                   0.023 &    \cellcolor{myred}19.4 \\
SE3+Focal+Intrinsics & \checkmark & & \checkmark            &                   28.13 & \cellcolor{myyellow}0.821 & \cellcolor{myyellow}0.203 &                   0.433 &                   0.022 &                   21.3 \\
SE3+Focal+Intrinsics & \checkmark & \checkmark & \checkmark &    \cellcolor{myred}28.20 & \cellcolor{myyellow}0.821 & \cellcolor{myorange}0.202 & \cellcolor{myyellow}0.429 &                   0.022 & \cellcolor{myyellow}21.1 \\
FocalPose+Intrinsics & \checkmark & & \checkmark            & \cellcolor{myyellow}28.18 & \cellcolor{myorange}0.822 &    \cellcolor{myred}0.201 & \cellcolor{myorange}0.427 &    \cellcolor{myred}0.016 &                   22.2 \\
\hline
FocalPose+Intrinsics & \checkmark & \checkmark & \checkmark & \cellcolor{myorange}28.19 &    \cellcolor{myred}0.823 &    \cellcolor{myred}0.201 &                   0.432 &    \cellcolor{myred}0.016 &                   21.7

\end{tabular}
}
\end{table}

We provide analysis on variant parameterizations focal length, camera-to-origin depth, and principal point.

\paragraph{\camerafont{Pixel Scale}} \camerafontnobf{Pix. Scale} scales pixel-valued parameters by the initial camera focal length $f$ (examples are focal length, principal point, and the translation parameters of \camerafont{FocalPose}). Multiplying by the focal length essentially scales the camera residual parameters such that they are encoded as world coordinates in the parameters. For example an additive focal length residual would be modified as
\begin{align}
    f' = f + f\Delta f \,,
\end{align}
the principal point residual are
\begin{align}
    u_0' = u_0 + f\Delta u_0 \\
    v_0' = v_0 + f\Delta v_0\,.
\end{align}

\paragraph{\camerafont{Log Scale}} Additive biases are not invariant to the initial absolute scale of a parameter. For example, if a camera and scene are simultaneously scaled up, the same bias will have a much smaller effect on the scene. For length-like quantities (focal-length, camera-to-object depth), it is therefore benefitial to parameterize them as a \emph{log scale}, as in \camerafont{FocalPose}.

\paragraph{Ablation}

In \tabref{tab:ablation_manual_reparam}, we show how these variants of our parameterizations affect performance. The table shows that careful consideration is required when choosing the parameterizations for a camera, and how preconditioning can help improve even sub-optimal parameterizations.

\end{anonsuppress}

\bibliographystyle{ACM-Reference-Format}
\bibliography{bibliography}

\end{document}